\documentclass[10pt,twocolumn,a4paper]{article}

\usepackage{times}
\usepackage{graphicx}
\usepackage{amsmath}
\usepackage{amssymb}
\usepackage{tabularx}
\usepackage{booktabs}
\usepackage{algorithmic}
\usepackage{algorithm}
\usepackage{textgreek}
\usepackage{makecell}
\usepackage{multirow}
\usepackage{dirtytalk}
\usepackage{biblatex}
\begin{document}

\title{D-DARTS: Distributed Differentiable Architecture Search}

\author{Alexandre Heuillet\\
Université Paris-Saclay\\
{\tt\small alexandre.heuillet@univ-evry.fr}
\and
Hedi Tabia\\
Université Paris-Saclay\\
{\tt\small hedi.tabia@univ-evry.fr}
\and
Hichem Arioui\\
Université Paris-Saclay\\
{\tt\small hichem.arioui@univ-evry.fr}
\and
Kamal Youcef-Toumi\\
MIT\\
{\tt\small youcef@mit.edu}
}

\date{}

\maketitle

\begin{abstract}
   Differentiable ARchiTecture Search (DARTS) is one of the most trending Neural Architecture Search (NAS) methods. It drastically reduces search cost by resorting to weight-sharing. However, it also dramatically reduces the search space, thus excluding potential promising architectures. In this article, we propose D-DARTS, a solution that addresses this problem by nesting neural networks at the cell level instead of using weight-sharing to produce more diversified and specialized architectures. Moreover, we introduce a novel algorithm that can derive deeper architectures from a few trained cells, increasing performance and saving computation time. In addition, we also present an alternative search space (DARTOpti) in which we optimize existing handcrafted architectures (e.g., ResNet) rather than starting from scratch. This approach is accompanied by a novel metric that measures the distance between architectures inside our custom search space. Our solution reaches competitive performance on multiple computer vision tasks.
\end{abstract}

\section{Introduction}

With the field of Deep Learning (DL) getting more and more attention over the past few years, a significant focus has been set on neural network architectural conception. A large number of architectures have been manually crafted to address different computer vision problems~\cite{he2016deep, howard2017mobilenets, simonyan15deep}. However, the design of these human-made architectures is mainly driven by intuition and lacks the certainty of an optimal solution. This is due to the vast number of possible combinations needed to build a relevant neural architecture, making the search space very difficult to browse manually. Neural Architecture Search (NAS) works~\cite{ren2020comprehensive, liu2019darts, chu2020fair, zoph2017neural} tried to tackle this issue by automatizing the architecture design process. In NAS, a search algorithm attempts to build a neural network architecture from a defined search space by asserting the performance of \say{candidate} architectures. The most trending NAS approach is currently Differentiable ARchitecTure Search (DARTS) \cite{liu2019darts} whose main advantage is a greatly reduced search cost compared to earlier approaches \cite{zoph2017neural, zoph2018learning} thanks to its 2-cell search space.

However, searching for only two types of cells significantly restricts the search space and limits the diversity of candidate architectures. To alleviate this problem, we propose to directly search for a complete network with individualized cells. This network delegates the search process to the cell level in a distributed fashion.
Moreover, we introduce a loss function based on Shapley values \cite{shapley_1953} to help optimize the super network architecture. Consequently, this new distributed supernet structure makes it possible to directly encode existing architectures (e.g., ResNet50~\cite{he2016deep}) in the search space and use them as starting points for the search process. Finally, we introduce a novel metric to compute distances between architectures in the search space, which can be considered a manifold.

As presented in Section~\ref{sec:approach}, the contributions of this article are:
\begin{itemize}
    \item A new way of structuring DARTS' search process by nesting small neural networks at the cell level.
    \item A novel loss specially designed to take advantage of the new distributed structure.
    \item A search space augmented with additional operations.
    \item A novel distance metric to highlight the structural differences between architectures.
\end{itemize}
The rest of the article is structured as follows: In Section~\ref{sec:related_work}, we conduct a short survey on recent differentiable NAS works, in Section~\ref{sec:darts}, we review the original concept of DARTS and discuss its issues, Section~\ref{sec:experiments} presents the results of a set of experiments conducted on popular computer vision datasets, Section~\ref{sec:discussion} discusses the results of the experimental study, and Section~\ref{sec:conclusion} brings a conclusion to this article while giving some insights on promising directions of future work.

\section{Related Work}
\label{sec:related_work}
In this section, we briefly review recent key studies related to our work. These include differentiable NAS, automatically improving existing handcrafted architectures, and quantifying the distance between architectures in a given search space.

\subsection*{Differentiable Neural Architecture Search}

Few other works already attempted to improve on DARTS by addressing its limitations (discussed in Section~\ref{sec:darts}). In PC-DARTS \cite{xu2020pcdarts}, the authors tried to minimize the memory footprint of DARTS by optimizing the search process to avoid redundancy. P-DARTS \cite{chen2021progressive} greatly reduced the search time by progressively deepening the architecture when searching, leading to a better search space approximation and regularization. FairDARTS \cite{chu2020fair} used the \textit{sigmoid} function to discretize architectures and introduced a novel loss function (see Section \ref{sec:darts}) leading to a \say{fairer} system. DARTS- \cite{chu2021darts} also strived to solve the critical issue of dominant \textit{skip} connections. To this end, they introduced an auxiliary skip connection with a $\beta$-decay factor. $\beta$-DARTS \cite{ye2022beta} improved DARTS- with a new regularization method (dubbed \textit{Beta-Decay}) that can prevent the architectural parameters from saturating. This led to increased robustness and better generalization ability. iDARTS \cite{zhang2021idarts} focused on improving the architecture gradient computation (\textit{hypergradient}). They introduced a stochastic hypergradient approximation for differentiable NAS based on the implicit function theorem. Finally, DOTS \cite{gu2021dots} proposed to decouple the operation and topology search, so that the cell topology is no longer constrained by the operation weights. Therefore, pairwise edge combinations are attributed weights that are updated separately from the operation weights. All of these works obtained competitive results on popular datasets~\cite{krizhevsky2012imagenet, krizhevsky2009learning}, but they nonetheless suffer from a restricted search space by searching only for two different type of cells {contrary to our D-DARTS approach (see Section \ref{sec:approach})}. Another work called FBNetV2 \cite{wan2020fbnetv2} does not rely on the DARTS process. It uses a masking mechanism for feature map reuse to increase the search speed and included a wider range of spatial and channel dimensions.
Motivated by the success of DARTS, this present work explores for the first time a gradient-based distributed search procedure for deep convolutional neural network architectures, thus addressing its restricted search space limitations.

\subsection*{Fine-Grained Neural Architecture Search}

Several previous articles strove to increase diversity in the candidate architectures by searching beyond the cell level. This approach is dubbed Fine-Grained Neural Architecture Search. Chaudhuri et al. \cite{Chaudhuri2020stochastic} managed to increase diversity in candidate architectures by searching for multiple blocks inside of cells and allowing operators inside of blocks to be modified during the search process. Lee et al. \cite{lee4006014fine} perform channel-wise neural architecture search using a differentiable channel pruning method. Their FGNAS model comprises several individual searchable residual blocks hosting multiple activation functions and inside which the channel pruning operation is done. Mei et al. \cite{mei2020atomnas} also performs neural architecture search at the channel level with independent cells. They introduced a supernet composed of very small \say{atomic} blocks comprising only a single mix of three operators (channel-wise operation) and progressively prune them to reduce the size of their final network. However, in contrast to these Fine-Grained NAS methods, our D-DARTS approach does not perform search beyond the cell level and instead focuses on making cells individual through a distributed process (see Section \ref{sec:approach}).

\subsection*{Automatically Improving Handcrafted Architectures}

Only a handful of studies tried to optimize handcrafted architectures through NAS. Luo et al.~\cite{luo2018neural} optimized a given architecture by constructing a performance predictor that maps the architecture's continuous representation with its performance. Guo et al.~\cite{guo2019nat} showcased a method called Neural Architecture Transformer that considers the problem as a Markov Decision Process (MDP) and makes use of Reinforcement Learning to automatically learn optimization policies for the different architectures.

\subsection*{Quantifying the Distance Between Architectures}

Although ample research was conducted on quantifying the distance between graphs~\cite{blondel2004measure, bouttier2003geodesic, sanfeliu1983distance}, very few studies considered the case of neural network architectures. Kandasamy et al.~\cite{kandasamy2018neural} developed a Bayesian Optimization framework for NAS which makes use of a distance metric called OTMANN. This metric is based on the number of computations performed in each layer of the two architectures whose distance is being asserted. Another work~\cite{hu2020angle} proposed to progressively shrink the search space. It uses an angle-based distance metric to guide this shrinking process.

\section{Preliminaries: DARTS}
\label{sec:darts}

Differentiable ARchitecTure Search (DARTS)~\cite{liu2019darts} is a gradient-based NAS method that searches for novel architectures through a cell-modulated search space while using a weight-sharing mechanism to speed up this process. More specifically, it searches for two different types of cells: \textit{normal} (i.e., that makes up most of the architecture) and \textit{reduction} (i.e., that performs dimension reduction). During the search process, a small size proxy network (supernet) with only a few of these cells (e.g., 8 as suggested in~\cite{liu2019darts}) is trained. These two cells are the building blocks from which architectures of any size can be derived, similar to the residual blocks in ResNet~\cite{he2016deep}. Thus, most of the final network components share the same architectural weights. In particular, the two cells are stacked multiple times to form a network of the desired size (e.g., 14 or 20 layers as in~\cite{liu2019darts}). This size is determined empirically depending on if the emphasis is put on raw performance or hardware efficiency. 

Each cell can be described as a direct acyclic graph of $N$ nodes where each edge connecting two nodes is a mix of operations chosen among $|O_{i,j}|=K$ candidates, where $O_{i,j} = \{o^{1}_{i,j}, ..., o^{K}_{i,j}\}$ represents the set of all possible operations for the edge $e_{i,j}$ connecting node $i$ to node $j$. DARTS browses a search space $S$ comprising $K=7$ operations (\textit{skip\_connect}, \textit{max\_pool\_3x3}, \textit{avg\_pool\_3x3}, \textit{sep\_conv\_3x3}, \textit{sep\_conv\_5x5}, \textit{dil\_conv\_3x3} and \textit{dil\_conv\_5x5}).

The goal of the DARTS process is to determine which operations (with a maximum of 2) must be selected for each edge in order to maximize the validation loss $L_{val}$. Such loss corresponds here to the Cross-Entropy loss $L_{CE}$, a widely used loss function derived from C.E. Shannon's Theory of Information \cite{shannon1948mathematical} and defined as follows:
\begin{equation}
\label{eq:cross_entropy}
    L_{CE}(x, t_c) = -log\left(\frac{exp(x[t_c])}{\sum_{j=0}^{|x|} exp(x[j])}\right) 
\end{equation}
where $x$ is the output of the final linear classifier (i.e. a probability distribution) and $t_c$ is the target class for this output.

In that objective, DARTS' search process involves learning a set of parameters, denoted by $\alpha_{i,j} = \{\alpha^{1}_{i,j}, ..., \alpha^{K}_{i,j}\}$, representing the weight of each operation from $O_{i,j}$ in the mixed output of each edge. To build the mixed output, the categorical choice of operations is done through a \textit{softmax}:

\begin{equation}
    \overline{o}_{i,j}(x) = \sum^{K}_{k=1} \frac{exp(\alpha^{k}_{i,j})}{\sum^{K}_{k'=1} exp(\alpha^{k'}_{i,j})}o^{k}_{i,j}(x)
\label{eq:softmax}
\end{equation}

where $\overline{o}_{i,j}(x)$ is the mixed output of edge $e_{i,j}$ for input feature $x$ and $\alpha^{k}_{i,j} \in \alpha_{i,j}$ is the weight associated with operation $o^k_{i,j} \in O_{i,j}$. 

\begin{figure*}[!ht]
    \centering
    \includegraphics[width=\linewidth]{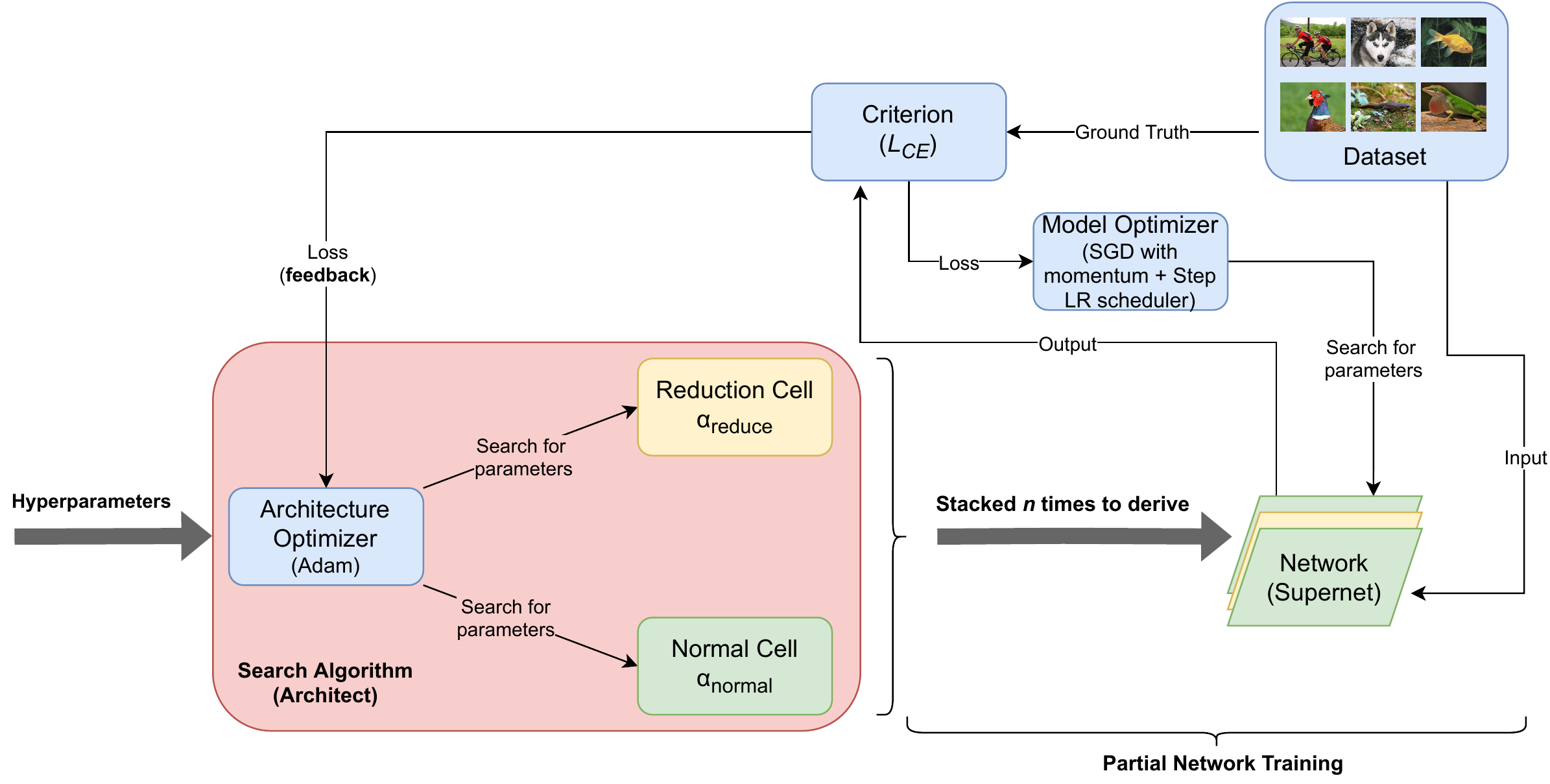}
    \caption{Layout of the search process at work in DARTS \cite{liu2019darts}. A global optimizer searches for two sets of parameters (i.e., $\alpha_{normal}$ and $\alpha_{reduce}$) that define the architecture of cells. The two types of searched cells are stacked multiple times to build a proxy network (supernet) trained to validate the performance of these cells.}
    \label{fig:darts}
\end{figure*}

These parameters are optimized using a gradient descent algorithm while the global supernet (from which the cells are part) is trained on a given dataset. Thus, DARTS is practically solving a bi-level optimization problem. The search process of DARTS is summarized in Fig.~\ref{fig:darts}.

However, DARTS suffers from two majors issues. The first is the over-representation of \textit{skip connections}. The second is the discretization discrepancy problem of the \textit{softmax} operation in Eq.~(\ref{eq:softmax}), namely a very small standard deviation of the resulting probability distribution. To mitigate these issues, the authors of FairDARTS~\cite{chu2020fair} replaced the \textit{softmax} function with the \textit{sigmoid} function (denoted by $\sigma$) :
\begin{equation}
    \overline{o}_{i,j}(x) = \sum^{K}_{k=1}\sigma(\alpha^k_{i,j})o^{k}_{i,j}(x) = \sum^{K}_{k=1}\frac{1}{1+exp(-\alpha^k_{i,j})}o^{k}_{i,j}(x). 
\end{equation}
They also proposed a novel loss function (\textit{zero-one loss} denoted by $L_{01}$) which aims to push the architectural weight values towards 0 or 1, and is defined for all the weights $\alpha$ as follows:
\begin{equation}
\label{eq:zero_one_loss}
    L_{01} = -\frac{1}{|\alpha|}\sum^{|\alpha|}_{i = 1}(\sigma(\alpha_i) - 0.5)^2, 
\end{equation}
where $|\alpha|$ is the total number of architectural weights in the cell.
In fact, $L_{01}$ corresponds to the mean square error between $\sigma(\alpha_i)$ and $0.5$.
$L_{01}$ is then added to $L_{CE}$, given by Eq.~(\ref{eq:cross_entropy}), to form FairDARTS total loss $L_F$:
\begin{equation}
\label{eq:fair_darts_loss}
    L_F = L_{CE} + {w_{0-1}L_{0-1}}
\end{equation}
where {$w_{0-1}$} is a coefficient weighting {$L_{0-1}$}.

Despite these solutions, DARTS and all of its evolutions \cite{chen2021progressive, chu2020fair, liu2019darts, xu2020pcdarts} are still limited in their capacity to create original architectures since most of their structure is rigid and human-made (e.g., the search space modulation or the number of cells searched). The search space is also very restricted as pointed out by prior works \cite{radosavovic2020designing, wan2020fbnetv2}, with only 2 types of cells, 14 edges per cell and 7 operations to choose from.

This is the issue we are trying to address in this article. To the best of our knowledge, our method is the first to explore and implement distributed differential neural architectural search.

\section{Proposed Approach}
\label{sec:approach}
In this section, we present the main contributions of our article: a novel distributed differentiable NAS approach, and a method to optimize existing handcrafted architectures using differentiable NAS. 

{To summarize, we nested neural networks inside cells to individualize them and designed a cell-specific loss to help with the optimization of this distributed process. Moreover, a supernet with individual cells is able to encode existing handcrafted architectures inside its search space. These architectures can then be used as starting points for the search process.}
\subsection{Delegating Search to Cell-Level Subnets}
\label{subsec:cell_search}
\begin{figure*}[!ht]
    \centering
    \includegraphics[width=\linewidth]{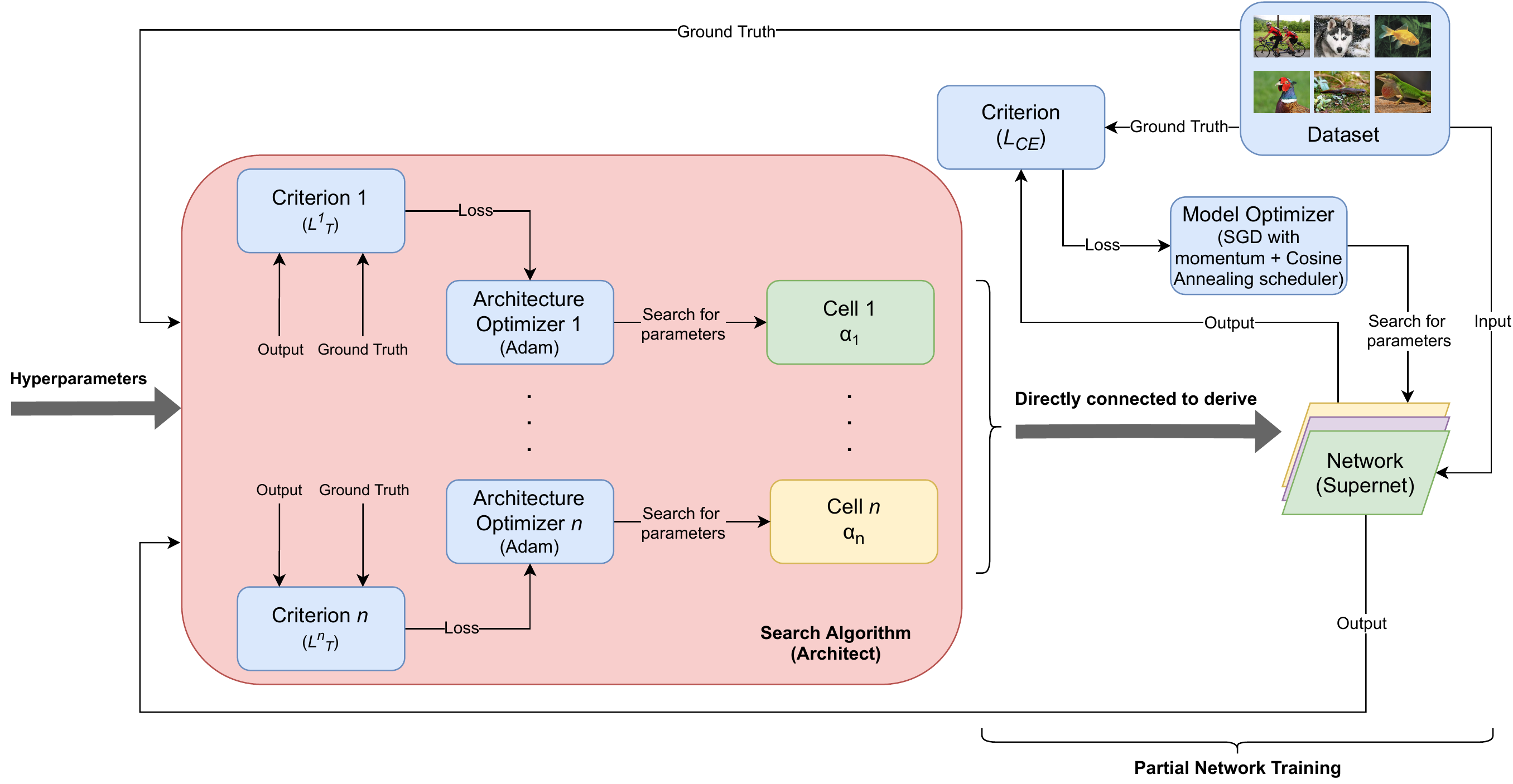}
    \caption{Layout of the search process used in D-DARTS. Cells are still divided into two types (\textit{normal}, \textit{reduction}) but each cell $i$ is independent with its own optimizer, scheduler and criterion (based on our novel ablation loss $L_{T}$). Each cell searches for its architectural parameters $\alpha_{i}$, which makes the entire search process distributed. The searched cells are directly connected to each other to build a proxy network (supernet) trained to validate their performance.}
    \label{fig:layout_nestedDARTS}
\end{figure*}

The key idea behind our method is to increase diversity in the architecture by delegating the search process to subnets nested in each cell. This way each cell that composes the global supernet is individual. It is itself a full neural network with its own optimizer, criterion, scheduler, input, hidden, and output layers, as shown in Fig.~\ref{fig:layout_nestedDARTS}. In addition, instead of searching for building blocks as DARTS~\cite{liu2019darts} do, we increase the number of searched cells to an arbitrary $n$ (e.g., 8 or 10) and directly seek for a full $n$-layer convolutional neural network. Thus, we trade the weight-sharing process introduced in DARTS for greater flexibility and creativity. Nonetheless, cells still belong either to the \textit{normal} or \textit{reduction} class, depending on their position in the network (\textit{reduction} cells are positioned at the $\frac{1}{3}$ and $\frac{2}{3}$ of the network).

The architectural parameters $\alpha$ are computed using logits. In this context, logits represent the probability distribution generated by the linear classifier placed at the end of the supernet. At each training step, they are computed using the global supernet and are passed down to cells that use them to compute their losses and gradients to update their architectural weights $\alpha$. This way, we directly build a complete neural network where each cell is highly specialized (w.r.t. its position in the supernet, contrary to generic building blocks).

As explained in section~\ref{sec:darts}, DARTS only searches for two types of cells (\textit{normal} and \textit{reduction}) and stacks them multiple times to form a network as deep as needed. This weight-sharing process has the advantage of reducing search space to a limited set of parameters (i.e., $\alpha_{normal}$ and $\alpha_{reduce}$), thus saving time and hardware resources. However, this approach limits both the search space size and the originality of the derived architectures as all the underlying structure is human-designed. In particular, the search space size $s(K, N)$ of a single cell with $K$ primitive operations to select from (within a maximum of 2 from different incoming edges) and $N$ steps (i.e., intermediary nodes) can be computed as follows:
\begin{equation}
\label{eq:search_space}
    s(K, N) = \prod_{i=1}^{N} \frac{(i+1)i}{2}K^2
\end{equation}
Following Eq. (\ref{eq:search_space}), using DARTS default parameters ($K = 7$ and $N = 4$), the search space of a single cell comprises around $10^{9}$ possible configurations. Thus, as both \textit{normal} and \textit{reduction} cells share the same $K$ and $N$, the total search space size of DARTS is around $10^{18}$ possibilities. This number is comparable to other differentiable NAS works \cite{cai2018proxylessnas, wan2020fbnetv2}, but far lower than those of Reinforcement Learning based NAS methods \cite{baker2016designing, zoph2017neural} that describe architecture topologies using sequential layer-wise operations, which are also far less efficient.

Our approach effectively expands the search space by a factor of $10^{(c-2)*9}$ (according to Eq.~(\ref{eq:search_space})) where $c$ is the total number of searched cells. Thus, the total size of D-DARTS search reaches around $10^{72}$ when considering $c = 8$. We dubbed this approach \textit{D-DARTS} for Distributed Differentiable ARchiTecture Search. In section~\ref{sec:experiments}, we show that smaller (e.g., 4 or 8 layers) D-DARTS architectures can achieve similar or higher performance than large (e.g., 14 or 20 layers) architectures on common computer vision datasets.

\subsection{Adding a New Cell-Specific Loss}
\label{subsec:loss}

In addition to the new network structure introduced in subsection~\ref{subsec:cell_search}, we designed a novel cell-specific loss function that we dubbed ablation loss. Indeed, as we increased the number of searched cells, the learning challenge became greater with a large amount of additional parameters to take into account. Thus, the global loss functions used in DARTS~\cite{liu2019darts} and FairDARTS~\cite{chu2020fair} cannot accurately assess the performance of each cell and instead only take into account the global performance of the supernet. In contrast, our new loss function is specific to each cell. It is an additive loss, based on the global loss function introduced in~\cite{chu2020fair} that proved to be a significant improvement over the original one~\cite{liu2019darts}.

The main idea behind this ablation loss function is to perform a limited ablation study on the cell level. This is done by measuring the performance of the network with/without each cell. Ablation studies have long proven to hold a key role in asserting the effectiveness of neural network architectures~\cite{meyes2019ablation}. This way, by computing the difference in the supernet loss $L_{CE}$ (i.e., the Cross-Entropy loss, see Eq.(~\ref{eq:cross_entropy})) with and without each individual cell activated, we can obtain a measure of their respective contributions that we call their marginal contributions, labeled $M_{C} = \{M_{C}^{1},...,M_{C}^{n}\}$ for an $n$-cell network. This method is inspired by Shapley values~\cite{shapley_1953}, a game theory technique widely used in Explainable Artificial Intelligence to assess the contributions of model features to the final output~\cite{ancona2019explaining, NIPS2017_7062} or the contributions of agents to the common reward in a cooperative multi-agent Reinforcement Learning context~\cite{heuillet2022explainability}. Thus, cell $C_{i}$ marginal contribution $M_{C}^{i}$ is computed as follows:

\begin{equation}
    M_{C}^{i} = L_{CE}^{(C)} - L_{CE}^{(C \setminus \{C_{i}\})},
\end{equation}
where $C$ is the set containing all cells such as $C = \{C_{1},...,C_{n}\}$.
Once we obtained all the marginal contributions $M_C$, we apply the following formula to compute the ablation loss of cell $C_{i}$:
\begin{equation}
    L_{AB}^{i} = 
        \begin{cases}
            \frac{M_{C}^{i} - mean(M_C)}{mean(M_C)} & \quad \text{if } mean(M_C) \neq 0 \\
            0 & \quad \text{else}
        \end{cases}
\end{equation}

$L_{AB}^{i}$ expresses how important the marginal contribution (i.e., its performance) of cell $i$ is w.r.t. the mean of all the marginal contributions. $L_{AB}^{i}$ is then added to FairDARTS global loss $L_F$ (see Eq.(~\ref{eq:fair_darts_loss})) to form the total loss $L_T$, weighted by the hyperparameter {$w_{AB}$}:
\begin{equation}
\label{eq:total_loss}
    L_T^{i} = L_F + {w_{AB}}L_{AB}^{i}
\end{equation}
Finally, when expanding Eq. (\ref{eq:total_loss}), we obtain:
\begin{equation}
    L_T^{i} = L_{CE} + {w_{0-1}L_{0-1}} + {w_{AB}}L_{AB}^{i}
\end{equation}

where $L_{CE}$ is the Cross-Entropy loss (see Eq.~(\ref{eq:cross_entropy})), $L_{01}$ is the Zero-One loss introduced in FairDARTS~\cite{chu2020fair} (see Eq.~(\ref{eq:zero_one_loss})) and {$w_{0-1}$} is an hyperparameter weighting {$L_{0-1}$}.


In section~\ref{sec:experiments}, we show that $L_T^{i}$ can warm start the search process and provide a substantial increase in performance. However, it also increases GPU memory usage, as shown in Section~\ref{sec:ablation_loss_impact}.

\subsection{Building Larger Networks from a Few Highly Specialized Cells} 

In previous works~\cite{chu2020fair, liu2019darts, xu2020pcdarts}, the final network architecture was derived from two searched cells (i.e., \textit{normal} and \textit{reduction}) which were stacked as much time as needed to build a network (e.g., 10, 15 or 20 cells).

However, as presented in subsection~\ref{subsec:cell_search}, in D-DARTS, we directly search for a \say{full} network of multiple individual cells instead of searching for building block cells as in DARTS~\cite{liu2019darts}. But, the downside of this method is that searching for a high number of cells is memory hungry (see Section~\ref{sec:memory_efficiency}), as each cell must possess its own optimizer, criterion, and parameters. This is not a critical issue as we show in section~\ref{sec:experiments} that a few (e.g., 4 or 8) of these highly specialized cells can outperform a large number of base cells. 

Nonetheless, it may be useful to use a larger number of cells without spending additional search time, especially when dealing with highly complex datasets such as ImageNet~\cite{imagenet_cvpr09}. Thus, to save computation time and still profit from a high number of cells, we developed a new algorithm to derive larger architectures from an already searched smaller one, inspired by what is done in DARTS~\cite{liu2019darts}. The key idea behind this concept is to keep the global layout of the smaller architecture with the reduction cells positioned at the $1/3$ and $2/3$ of the network, similarly as in DARTS and FairDARTS~\cite{chu2020fair}, and repeat the searched structure of \say{normal} cells in the intervals between the reduction cells until we obtain the desired number of cells. This process is summarized in Algorithm~\ref{algo:arch_deriv}.

\begin{algorithm}[ht!]
\caption{Algorithm describing the larger architecture derivation process for D-DARTS}
\label{algo:arch_deriv}
\begin{algorithmic}[1]
\REQUIRE List: $C$, list of searched cells
\REQUIRE Integer: $n$, desired number of cells
\ENSURE List: $C_{f}$, list of cells that compose the derived architecture
\STATE $C_{f} \gets \text{empty\_list()}$
\STATE $m \gets \text{euclidean\_division}(|C|, 3)$
\STATE $m2 \gets \text{euclidean\_division}(2\times|C|, 3)$
\FOR{$i$ in $[0, n]$}
    \IF{$n > |C|$}
        \IF{$i < \text{euclidean\_division}(n, 3)$}
            \STATE $c \gets \text{modulo}(i,m)$
        \ELSIF{$i = \text{euclidean\_division}(n, 3)$}
            \STATE $c \gets m$
        \ELSIF{$i > \text{euclidean\_division}(n, 3)$ \AND $i < \text{euclidean\_division}(2\times n, 3)$}
            \STATE $c \gets \text{modulo}(i,m2-1-m) + m + 1$
        \ELSIF{$i = \text{euclidean\_division}(2\times n, 3)$}
            \STATE $c \gets m2$
        \ELSE
            \STATE $c \gets \text{modulo}(i,|C|-1-m2) + m2 + 1$
        \ENDIF
    \ELSE
        \STATE $c \gets i$
    \ENDIF
    \STATE $\text{append}(c, C_{f})$
\ENDFOR
\end{algorithmic}
\end{algorithm}

Thus, Algorithm~\ref{algo:arch_deriv} allows us to obtain a larger architecture without launching a new search (i.e., without any overhead).
In section~\ref{sec:experiments}, we show that doubling the number of layers this way can result in a 1\% increase of top1 accuracy when evaluating on CIFAR-100~\cite{krizhevsky2009learning}. However, the gain is more limited (around 0.15 \%) for simpler datasets such as CIFAR-10~\cite{krizhevsky2009learning}, where the base model already performs very well.

\subsection{Encoding Handcrafted Architectures in DARTS}
\label{sec:encoding_darts}

The use of D-DARTS distributed design allows the encoding of handcrafted architectures. This is one of the main advantages of the approach presented in Subsection \ref{subsec:cell_search}. Unlike the original DARTS \cite{liu2019darts} or one of its derivatives (such as P-DARTS \cite{chen2021progressive} or PC-DARTS \cite{xu2020pcdarts}), this method individualizes each cell and makes it possible to encode large handcrafted architectures \cite{chollet2017xception, he2016deep, szegedy2017inception}, which typically possess multiple types of layers (e.g., 5 for InceptionV4 \cite{szegedy2017inception} or 13 for Xception \cite{chollet2017xception}). Naturally, this would not have been possible using DARTS classic search space, which is limited to only two different cells. In addition, the primary motivation behind the encoding of handcrafted architectures is to use them as initial points for the optimization process. Since these traditional handcrafted architectures have carefully been optimized, they might be considered as local minima of the search space.

This process involves a few key procedures. These procedures are (i) encoding the handcrafted architecture in D-DARTS' cell system, (ii) a new weight-sharing mechanism, (iii) pretraining the architecture, and (iv) designing a new search space $S_o$.

(i) The architecture is manually encoded as a D-DARTS-compatible \textit{genotype} (i.e., a data structure that describes each cell design, the location of reduction cells in the architecture, and the maximum number of steps in a cell). Then, when searching from this architecture or training it, the corresponding \textit{genotype} is automatically loaded and deserialized into $\alpha$ weights (see Section \ref{sec:darts}), which can be optimized by D-DARTS' search process.

(ii) A weight-sharing mechanism is introduced to reduce the search cost and redundancy in cells (especially when starting from architectures with a large number of layers). Every identical cell in the baseline architecture will share the same weights. For instance, Xception \cite{chollet2017xception} is composed of 13 cells but only 5 of those are different. Thus, in this case, with our proposed weight-sharing mechanism, the number of optimizers is reduced from 13 to 5.

(iii) The supernet is pre-trained for 5 epochs. This happens before the actual search starts for the performance of the baseline architecture to be assessed and taken into account by the search algorithm.

(iv) 5 new operations are added to DARTS' 7 original ones (i.e., search space $S$) in order to implement the 3 handcrafted architectures described in Section \ref{sec:experiments} (i.e., ResNet18 \cite{he2016deep}, ResNet50 \cite{he2016deep} and Xception \cite{chollet2017xception}). These are (1) \texttt{conv\_3x1\_1x3}, (2) \texttt{conv\_7x1\_1x7}, (3) \texttt{simple\_conv\_1x1}, (4) \texttt{simple\_conv\_3x3}, and (5) \texttt{bottleneck\_1x3x1}. This brings the total number of operations to 12, and unlocks new possibilities but at the cost of further increasing D-DARTS' already large search space. This new search space is denoted $S_o$.

\begin{figure*}
    \centering
    \includegraphics[width=\linewidth]{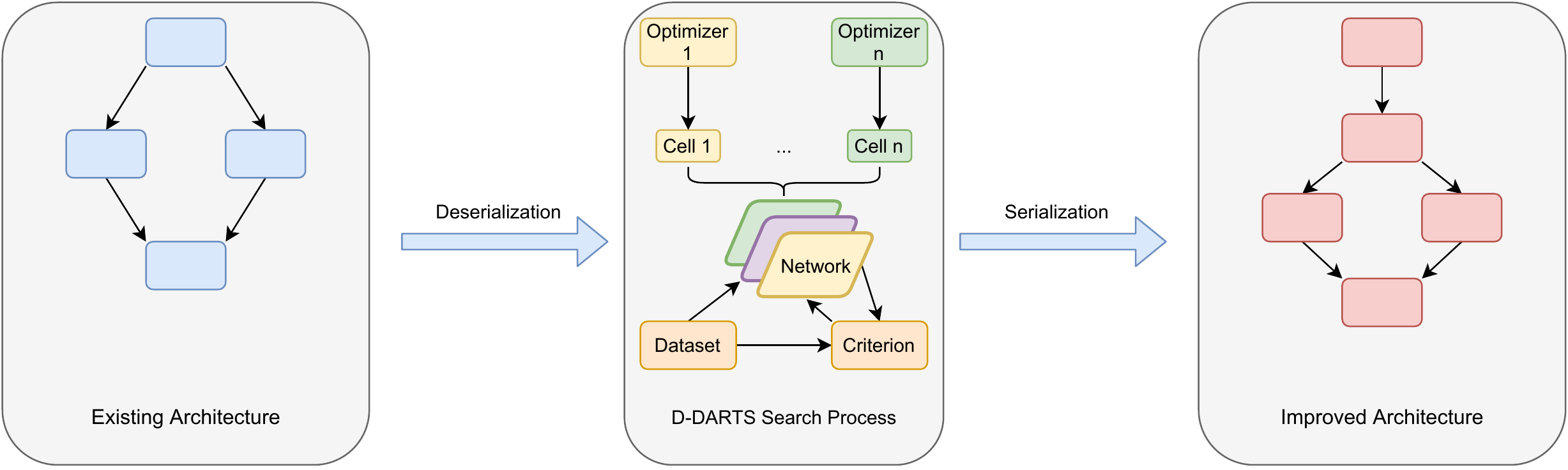}
    \caption{Overall description of the process used in DARTOpti.}
    \label{fig:DARTOpti_process}
\end{figure*}

This process, denoted DARTOpti, is visually summarized in Figure~\ref{fig:DARTOpti_process}. We show in Section~\ref{sec:experiments} that it can successfully optimize handcrafted architectures on ImageNet~\cite{krizhevsky2012imagenet}.

\subsection{Implementing a Metric to Quantify the Distance Between Architectures}
\label{sec:distance_metric}
We propose a metric $M$ able to effectively assert the distance between two architectures that belong to the search space $S_o$. This was motivated by the need to quantify how much DARTOpti (see Section \ref{sec:encoding_darts}) improved handcrafted architectures. Another motivation for $M$ is the possibility to draw statistics on $S_o$ such as the average distance between starting and final architectures, the maximum distance between starting and final architectures, or finding if the amelioration of performance in the final architectures is correlated to the number of changes made to the original ones. 


The metric $M$ is defined as follows. First, we need to compare (i) architectural operations with each other, then  (ii) cell edges, (iii) cells, and finally (iv) architectures. 

(i) A way to compare operations is needed, as they are the most elementary components of a \textit{genotype}. Naturally, we cannot consider all operations as having an equal value. They are all different, with some seeming to share more similarities. For instance, one would think that \texttt{sep\_conv\_3x3} has more in common with the other convolution operations than with the pooling operations. To that end, we established an experimental protocol that would allow us to assign a performance score to each operation, assuming that similar operations would also have similar scores. In particular, each of the 12 operations in $S_o$ (see Subsection~\ref{sec:encoding_darts}) was benchmarked for each edge of each cell of a small 3-cell proxy network {based on ResNet18 \cite{he2016deep} architecture (i.e., residual blocks). This way, the proxy network is similar to an existing CNN architecture}. All edge operations were disabled except for the benchmarked one. We recorded the maximum top-1 accuracy reached by the proxy network among 4 training runs of 10 epochs each (to counteract the stochasticity of the gradient descent algorithm) on the CIFAR-10~\cite{krizhevsky2009learning} dataset. {Training the proxy network took 1 GPU-day on a single Nvidia RTX A6000 with a batch size of 256.} Final scores for each operation were obtained by taking the median value observed across all edges of all cells. Results are presented in Table~\ref{tab:op_scores}. They conform to expectations with analogous operations obtaining comparable scores (e.g., \texttt{dil\_conv\_3x3} and \texttt{dil\_conv\_5x5} are only 0.03 \% apart).

\begin{table}[ht]
    \centering
    \caption{Benchmark scores obtained for each of the 12 operations in DARTOpti search space $S_o$. Standard deviation across all cell edges is reported for each operation.}
    \begin{tabular}{cc}
        \toprule
        Operation & Score (in \%) \\
        \midrule
        \texttt{conv\_3x1\_1x3} & $82.76 \pm 1.82$ \\
        \texttt{conv\_7x1\_1x7} & $82.72 \pm 1.14$ \\
        \texttt{max\_pool\_3x3} & $82.96 \pm 2.19$ \\
        \texttt{avg\_pool\_3x3} & $82.51 \pm 2.13$ \\
        \texttt{skip\_connect} & $82.15 \pm 2.33$ \\
        \texttt{simple\_conv\_1x1} & $82.27 \pm 1.99$ \\
        \texttt{simple\_conv\_3x3} & $83.12 \pm 2.21$ \\
        \texttt{sep\_conv\_3x3} & $83.19 \pm 0.89$ \\
        \texttt{sep\_conv\_5x5} & $84.87 \pm 0.73$ \\
        \texttt{dil\_conv\_3x3} & $82.96 \pm 1.56$ \\
        \texttt{dil\_conv\_5x5} & $82.99 \pm 1.24$ \\
        \texttt{bottleneck\_1x3x1} & $83.06 \pm 0.77$ \\
        \bottomrule
    \end{tabular}
    \label{tab:op_scores}
\end{table}

(ii) The distance between two edges, each encoded as a binary vector, can be computed using the Hamming distance \cite{hamming1950error}. This widely used function measures the distance between vectors by evaluating the minimum number of changes needed to translate one vector into the other according to given weights. Here, these weights correspond to the operation score values featured in Table~\ref{tab:op_scores}. The Hamming distance is described in Algorithm \ref{algo:hamming}.

\begin{algorithm}[ht!]
\caption{Algorithm describing the Hamming distance}
\label{algo:hamming}
\begin{algorithmic}[1]
\REQUIRE List: $U$, first vector to compare
\REQUIRE List: $V$, second vector to compare
\REQUIRE List: $W$, weights associated with vector indices
\STATE $n \gets |U|$
\STATE $dist \gets \text{empty\_list()}$
\FOR{$i$ in $[0, n]$}
    \IF{$U[i] != V[i]$}
        \STATE $dist[i] \gets W[i]$
    \ELSE
        \STATE $dist[i] \gets 0$
    \ENDIF
\ENDFOR
\RETURN $mean(dist)$
\end{algorithmic}
\end{algorithm}

(iii) The Hausdorff distance~\cite{hausdorff1914} is used to evaluate the distance between cells. This is described in the following equation:
\begin{equation}
\label{eq:hausdorff_distance}
    d_H(X,Y) = \text{max}\{\text{sup}_{x \in X}d(x, Y), \text{sup}_{y \in Y}d(X, y)\}  
\end{equation}
where $X$ and $Y$ are two subsets of $S_o$ (i.e., cells in this case) and $d$ is a distance metric able to quantify the distance between points of the subsets. Here, $d$ corresponds to the Hamming distance. The Hausdorff distance is relative to the closeness of every point of one set to those of the other set. Thus, it better considers structural similarity, which is paramount for comparing topologies. However, since the Hausdorff distance of Eq. (\ref{eq:hausdorff_distance}) is directive, we need to enforce symmetry in order to create a metric $M_H$:
\begin{equation}
    M_H(X,Y) = \text{max}(d_H(X,Y), d_H(Y,X))
\end{equation}

(iv) We can compare two different architectures $A$ and $B$ by evaluating the average distance between their pairwise cells with the architectural distance metric $M$:
\begin{equation}
    M(A,B) = \sum_{i=0}^{n} \frac{M_{H}(C_{i}^{A}, C_{i}^{B})}{n}
    \label{eq:distance_metric}
\end{equation}
where $n$ is the number of cells in both architectures. In fact, since every cell is individual and linearly linked to the others, it is not relevant to compare cells that are not in the same position in both architectures.
Moreover, as $M$ is a composition of metrics (Hausdorff, Hamming), it is a metric itself and so satisfies the basic properties of metric functions:
\begin{itemize}
    \item Identity of indiscernibles: $M(X, Y) = 0 \Leftrightarrow X = Y$
    \item Symmetry: $M(X, Y) = M(Y, X)$
    \item Triangle inequality: $M(X, Y) \leq M(X, Z) + M(Y, Z)$
\end{itemize}
In Subsection \ref{subsec:leveraging_metric} we show that, in addition to simply comparing two architectures, $M$ can also be used to find the optimal number of epochs for the search process, thus actively reducing search cost.

\section{Experiments}
\label{sec:experiments}

In this section, we conducted {image classification, object detection, and instance segmentation} experiments on several popular datasets {such as CIFAR \cite{krizhevsky2009learning}, ImageNet \cite{imagenet_cvpr09}, MS-COCO \cite{lin2014microsoft}, and Cityscapes \cite{Cordts2016Cityscapes}}. We also took advantage of the architectural distance metric $M$ presented in Subsection \ref{sec:distance_metric} to gain information about the search process and draw statistics from the search space.

\subsection{Experimental Settings}
\label{sec:experimental_settings}
{Image classification tasks were evaluated on CIFAR-10, CIFAR-100~\cite{krizhevsky2009learning} and ImageNet~\cite{krizhevsky2012imagenet} datasets while object detection and instance segmentation tasks were evaluated on MS-COCO \cite{lin2014microsoft} and Cityscapes \cite{Cordts2016Cityscapes} datasets respectively. All experiments were conducted using Nvidia GeForce RTX 3090 and Tesla V100 GPUs.}

We mostly use the same data processing, hyperparameters and training tricks as in FairDARTS~\cite{chu2020fair} and DARTS~\cite{liu2019darts} (e.g., AutoAugment~\cite{cubuk2019autoaugment} or label smoothing). We searched for 8-cell networks and used Algorithm \ref{algo:arch_deriv} to derive 14-cell networks.

Moreover, as discussed in Subsection \ref{sec:memory_efficiency}, the new search algorithm and loss increase GPU memory requirements. However, this has been alleviated by using Automatic Mixed Precision (AMP) to speed up floating-point operations and save video memory by working on half-size tensors. Thus, the freed memory allowed us to set batch size to 128 when searching on CIFAR datasets and to 96 when searching on ImageNet.

In addition, we assessed the consistency of the optimization technique presented in Subsection \ref{sec:encoding_darts} (DARTOpti). We used top-performing convolutional architectures as starting points for the search process: ResNet18 \cite{he2016deep}, ResNet50 \cite{he2016deep}, and Xception \cite{chollet2017xception}. In this case, the number of initial channels is increased to 64 when training to match the designs of the baseline architectures. However, this makes the average number of parameters of DARTOpti architectures significantly more important than in D-DARTS' ones (see Tables \ref{tab:cifar10}, \ref{tab:cifar100} and \ref{tab:imagenet}). Hence, we reduce the number of channels to 32 while searching to save video memory. Naturally, we searched for networks whose number of cells matches the number of different layers in the original architecture (e.g., 4 in ResNet50~\cite{he2016deep}). 

Finally, we select the architectural operations using FairDARTS \textit{edge} (i.e., 2 operations maximum per edge) or \textit{sparse} (i.e., 1 operation maximum per edge) method with a threshold of $0.85$ as in~\cite{chu2020fair}. We chose $w_{01}=8$ and $w_{abl}=0.5$ for the hyperparameters of total loss $L_{T}$ (see Eq.~(\ref{eq:total_loss})) as discussed in Section~\ref{sec:ablation_loss_impact}. DARTS, on the other hand, systematically selects the two operations with the highest \textit{softmax} weights for each edge (this parsing method is referred to as \textit{darts} in Table~\ref{tab:cifar10} and Table~\ref{tab:imagenet}). Our implementation is based on PyTorch 1.10.2 and is derived from~\cite{liu2019darts, chu2020fair}. Code and pretrained models can be accessed at \url{https://github.com/aheuillet/D-DARTS}.

\subsection{Analysis of Ablation Loss: What Impact Does it Have?}
\label{sec:ablation_loss_impact}

\subsubsection{Hyperparameter Choice}
We made the hyperparameter weights of the $L_{AB}$ ablation loss $w_{abl}$ and the $L_{01}$ zero-one loss $w_{01}$ from Eq.~(\ref{eq:total_loss}) vary to choose their optimal value w.r.t. the global loss. 

\begin{figure}[ht]
    \centering
    \includegraphics[width=\linewidth]{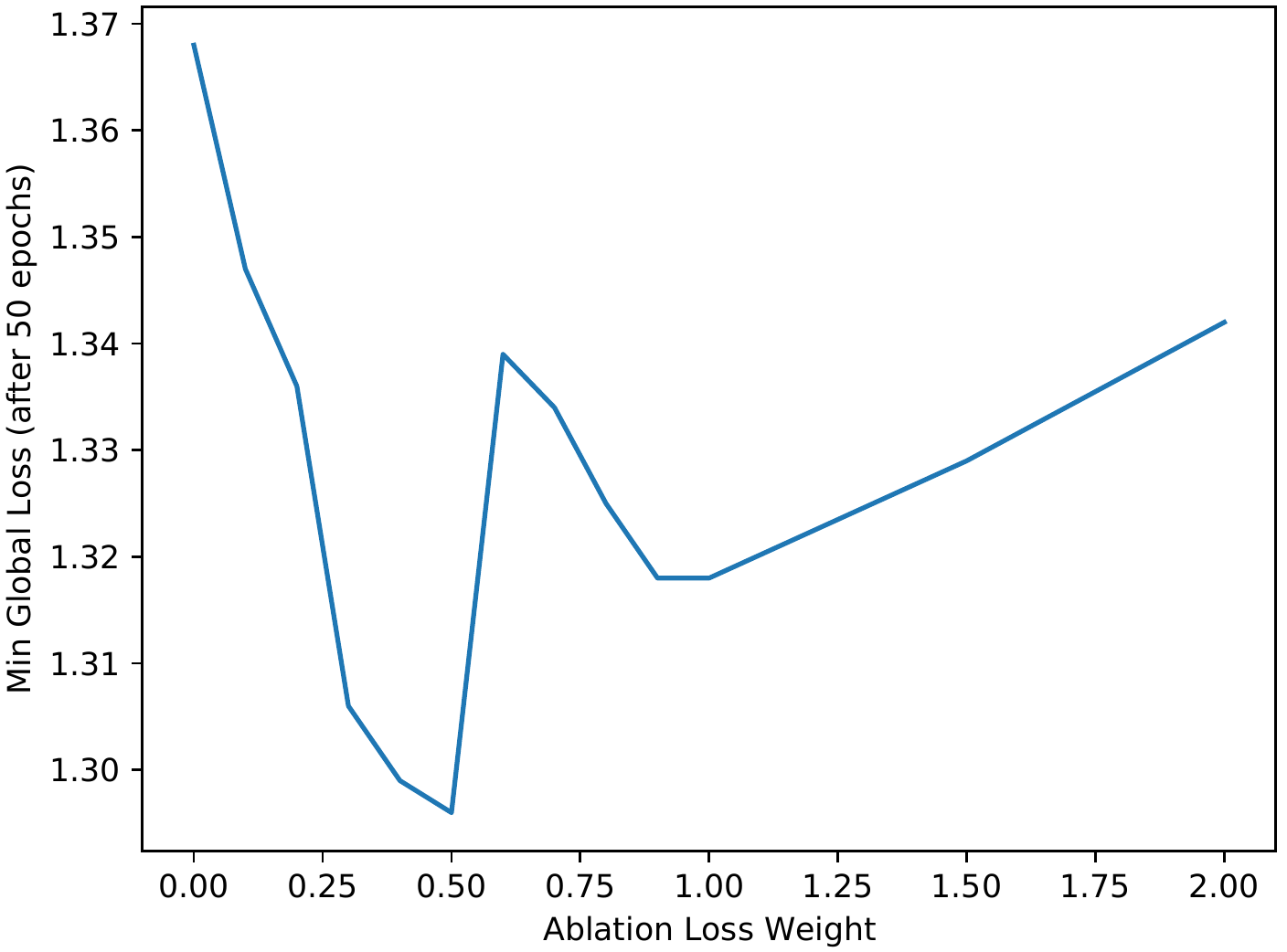}
    \caption{Line plot of the minimal global loss obtained by searching for a model on CIFAR-100 \cite{krizhevsky2009learning} for 50 epochs w.r.t. the sensitivity weight $w_{abl}$ used for the ablation loss $L_{AB}$. We deactivated $L_{01}$ (i.e., we set $w_{01}=0$) to prevent interference from occurring.}
    \label{fig:abl_min_glob_loss}
\end{figure}

Thus, in Fig.~\ref{fig:abl_min_glob_loss} we made $w_{abl}$ vary from 0 to 2 while keeping $L_{01}$ deactivated (i.e., $w_{01}=0$) in order to analyze its impact. We can observe that an optimal value seems to be attained around 0.5, with the global loss mainly increasing when $w_{abl}$ reaches higher or lower values.

\begin{figure}[ht]
    \centering
    \includegraphics[width=\linewidth]{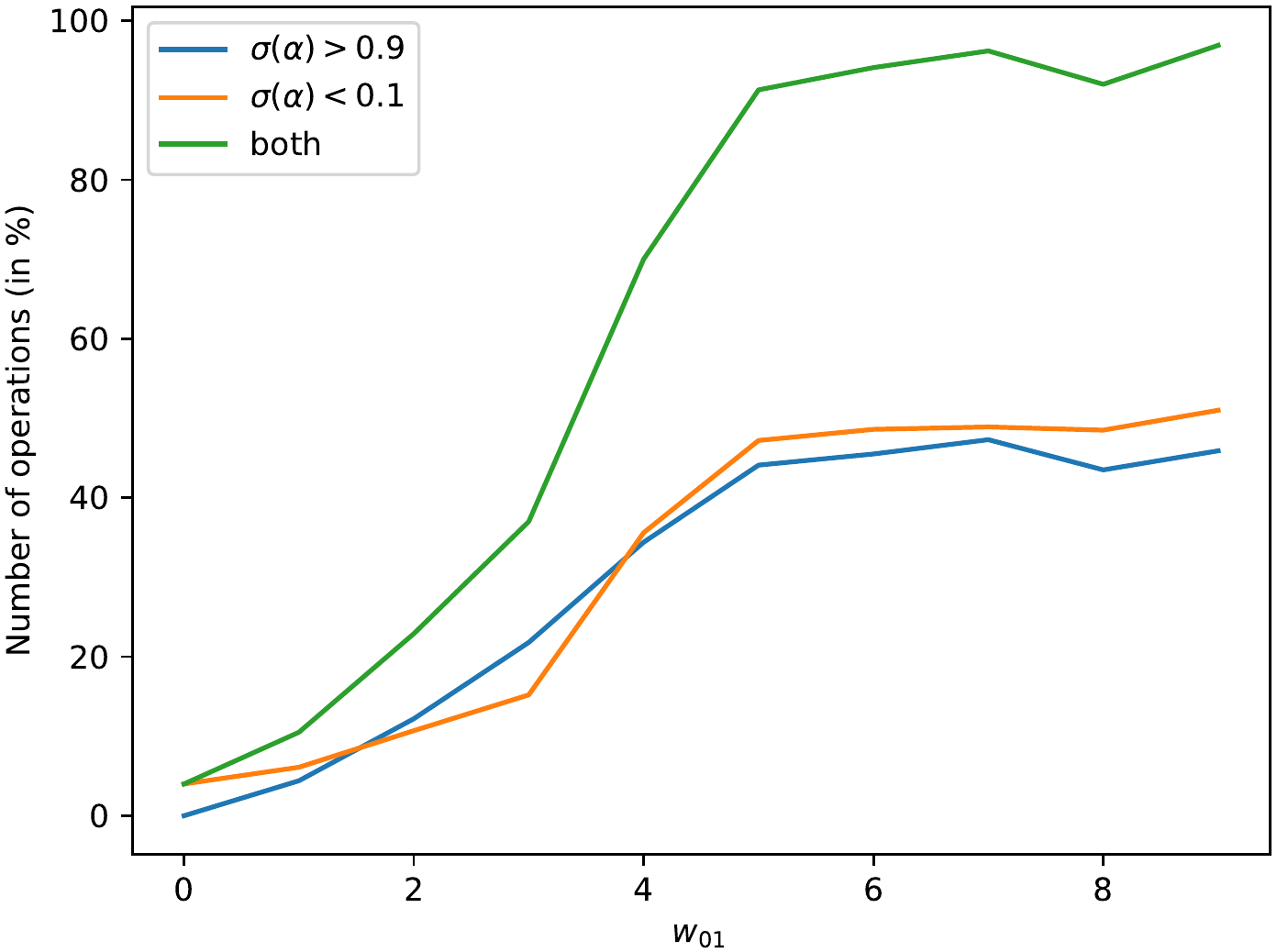}
    \caption{Line plot showing the percentage of dominant operations obtained in the final architecture $\alpha$ while searching on CIFAR-100 w.r.t. the sensitivity weight $w_{01}$ used for $L_{01}$. We can see that the proportion of both types of operations stabilizes after $w_{01}=5$ and reaches an equilibrium at $w_{01}=7$.}
    \label{fig:fair_loss}
\end{figure}

Moreover, we made the value of $w_{01}$ vary during search on CIFAR-100, with $w_{abl}=0.5$ fixed, and reported the number of dominant operations (i.e., operations whose sigmoid weight value $\sigma(\alpha)$ is greater than 0.9). This experiment was conducted to select a relevant value for $w_{01}$ since the one chosen by the authors of FairDARTS \cite{chu2020fair} ($w_{01}=10$) is no longer valid as we altered the search process. Fig. \ref{fig:fair_loss} shows that the proportion of dominant operations steadily increases from $w_{01}=0$ to $w_{01}=5$ where it reaches a plateau and stabilizes. It is worth noting that for $w_{01}=5$ and higher, nearly all sigmoid values $\sigma(\alpha)$ are either greater than 0.9 or inferior to 0.1. Finally, we chose 7 as the optimal value for $w_{01}$, as it offers both a high number of dominant operations and an equilibrium between operations with $\sigma(\alpha)>0.9$ and those with $\sigma(\alpha)<0.1$.

\subsubsection{Ablation Study}
We conducted an ablation study on our proposed ablation loss $L_T$ of Eq. (\ref{eq:total_loss}). In particular, we compared the performance of architectures with similar characteristics searched either with $L_T$ or FairDARTS \cite{chu2020fair} loss $L_F$ (see Eq. (\ref{eq:fair_darts_loss})). Tables~\ref{tab:cifar10} and~\ref{tab:cifar100} show that $L_T$-searched architectures (DD-1, DD-3, DD-4, DD-5) outperform their $L_F$-searched counterparts by an average of 0.6 \% across all datasets, confirming the advantage procured by this new loss function.

Concretely, when considering the 14-cell \textit{edge} parsed models evaluated on CIFAR-10 (DD-2 and DD-3), DD-3 reached a top-1 accuracy of 97.58 \%, thus outperforming $L_F$-searched DD-2 by 0.48 \%. Moreover, it is worth noting that $L_T$-searched DD-1 (8-cell model) reached a similar score as DD-2 (around 97.1 \%), despite featuring significantly fewer parameters (1.7M vs. 3.3M). One additional point is that $L_T$ seems to provide a larger increase in performance for CIFAR-100. For example, the gain in performance is around 1 \% between the 8-cell versions of DD-6 and DD-4.

However, when considering models that leveraged Algorithm \ref{algo:arch_deriv} to increase their number of cells (e.g., DD-4 and DD-6), the performance gain is slightly significant (e.g., around $0.1$ \% on CIFAR-100). This could be due to Algorithm \ref{algo:arch_deriv} that may have a leveling effect by disturbing the cell sequence and increasing the number of model parameters.

\subsubsection{Convergence Speed}
\begin{figure}
    \centering
    \includegraphics[width=\linewidth]{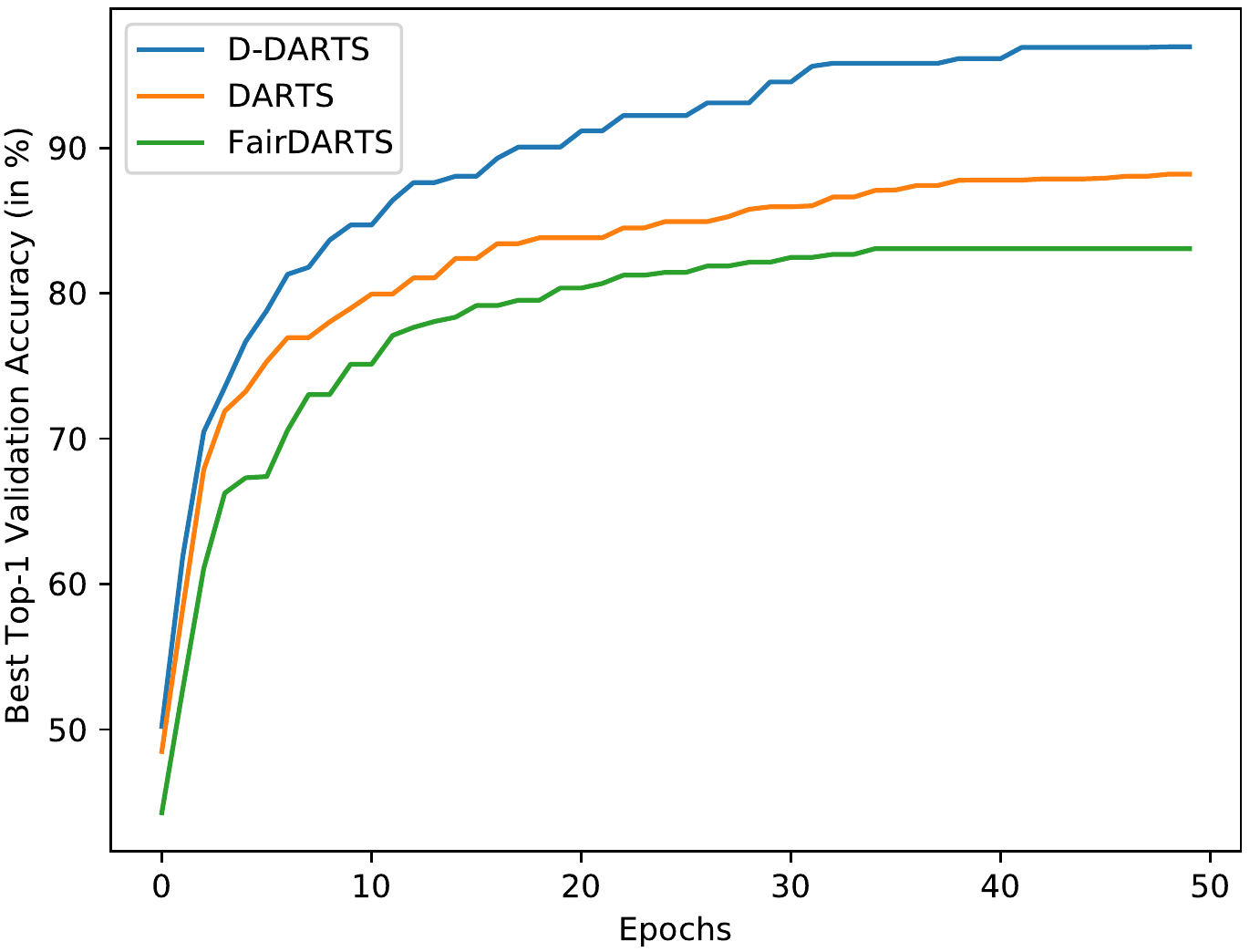}
    \caption{Line plot showing the best validation top-1 accuracy while searching on CIFAR-10\cite{krizhevsky2009learning} w.r.t. the current epoch. D-DARTS clearly outperforms both DARTS and FairDARTS by a large margin.}
    \label{fig:convergence_speed}
\end{figure}

We conducted an experiment on the search process convergence speed of our method D-DARTS compared to previous baselines~\cite{liu2019darts, chu2020fair}.
Fig.~\ref{fig:convergence_speed} shows a plot of the best validation top-1 accuracy w.r.t. the number of epochs. One can notice that D-DARTS converges very quickly and faster than the others. It hits a final plateau around epoch $40$. D-DARTS outperforms both DARTS and FairDARTS respectively by $9\%$ and $14\%$.

\subsection{Memory Efficiency}
\label{sec:memory_efficiency}
When searching using $L_T$ (see Eq. (\ref{eq:total_loss})), additional tensors must be stored on GPU memory due to the computations required to obtain the marginal contribution of each cell. The memory usage with $L_T$ is thus higher than what an $L_F$-based search requires. A $L_T$ search may increase the memory consumption by 100\%. For example, using an Nvidia RTX 3090 on CIFAR-100~\cite{krizhevsky2009learning} with a batch size of 72, the memory usage increases from $11100\text{ MB}$ with $L_F$ to $21911\text{ MB}$ of memory with $L_T$. Consequently, this can be identified as an issue of our method: we have to find means to reduce this memory consumption to search for deeper networks and make our method available to lower-end GPUs. 

Thus, it motivated the {usage} of an optimization technique: AMP (as presented in Section \ref{sec:experimental_settings}). AMP automatically converts tensors to half-precision (16 bits floats) when full precision (32 bits floats) is not required. {It is directly available in PyTorch starting from version 1.5.} Enabling this functionality effectively reduced memory consumption to 13000 MB when using $L_{T}$. Hence, AMP offers more flexibility to D-DARTS to run on GPUs equipped with less video memory and increase the batch size to achieve better memory utilization.

\subsection{Leveraging the Architectural Distance Metric}
\label{subsec:leveraging_metric}
\begin{figure}[ht]
    \centering
    \includegraphics[width=\linewidth]{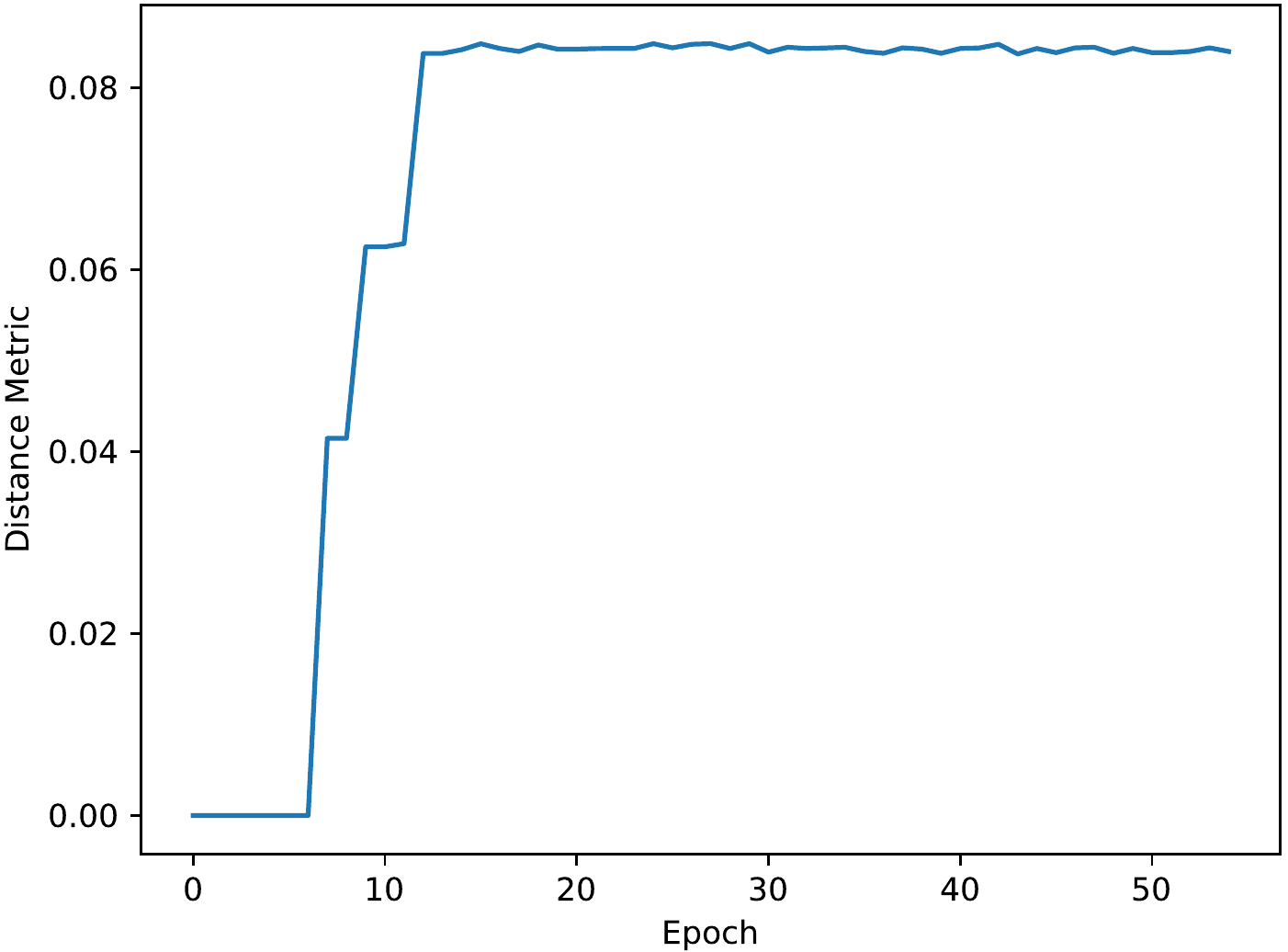}
    \caption{Line plot of the distance metric between the original ResNet18 \cite{he2016deep} architecture and the one being optimized by DARTOpti on CIFAR10 \cite{krizhevsky2009learning} according to the current epoch. The distance quickly rises to around 0.084 DU (distance units) at epoch 15 and then stabilizes. This indicates that no additional major changes are applied to the architecture after this point.}
    \label{fig:distance_metric_resnet18}
\end{figure}

\begin{figure}[ht]
    \centering
    \includegraphics[width=\linewidth]{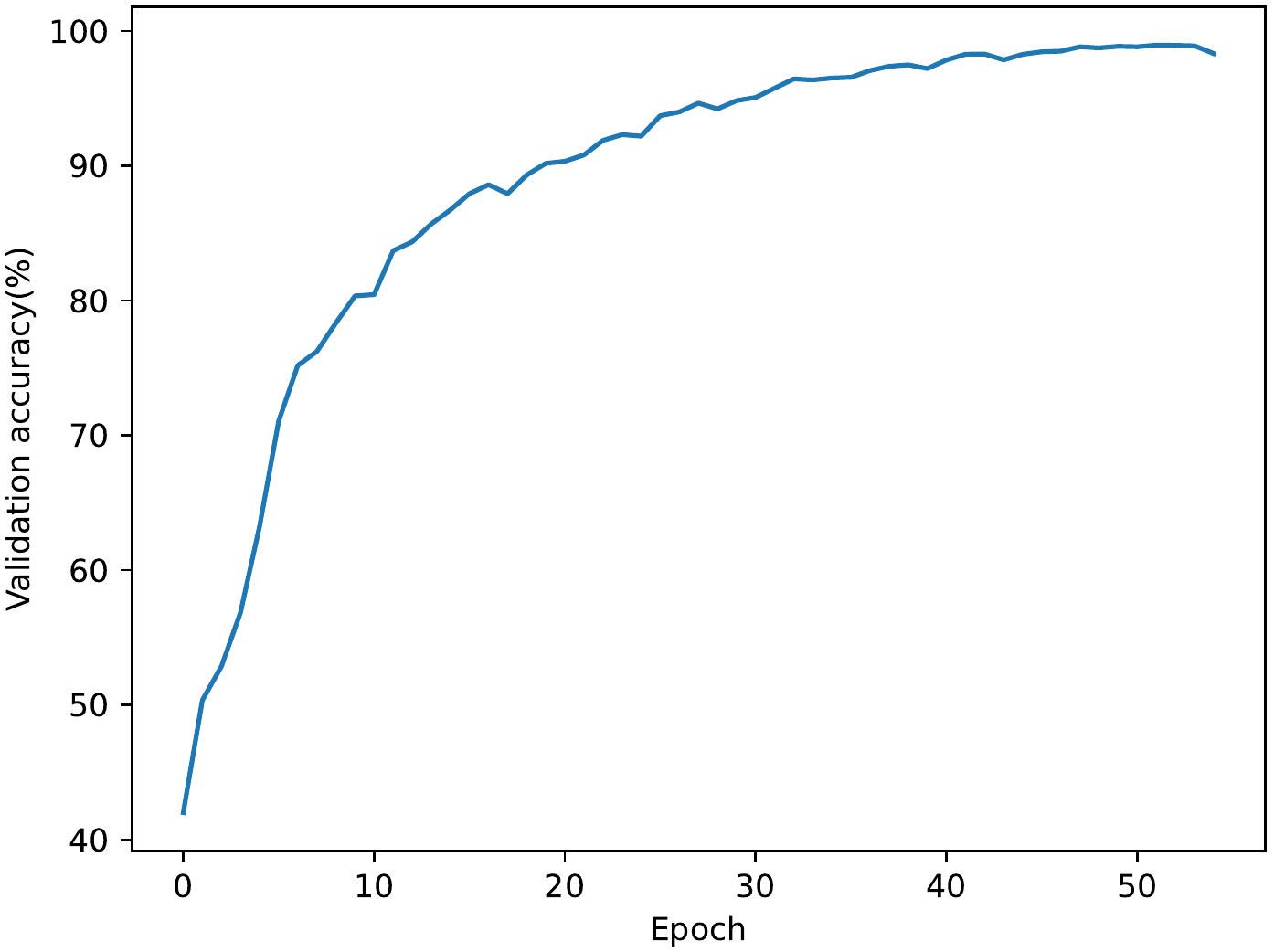}
    \caption{Line plot of the validation accuracy reached at each epoch on CIFAR-10 \cite{krizhevsky2009learning} while optimizing ResNet18 \cite{he2016deep} with DARTOpti.}
    \label{fig:val_acc_cifar10}
\end{figure}

The architectural distance metric $M$ (given by Eq. (\ref{eq:distance_metric})) allowed us to gain insightful information on the model and search process. First, we plotted $M$ associated with the current epoch w.r.t. the original architecture while running the optimization process. This way, we obtained plots such as Fig. \ref{fig:distance_metric_resnet18}, featuring $M$ computed for ResNet18 \cite{he2016deep} optimized on CIFAR-10 \cite{krizhevsky2009learning}, and Fig. \ref{fig:val_acc_cifar10}, presenting the validation accuracy reached by the same model during search. In Fig. \ref{fig:distance_metric_resnet18}, we can see that $M$ quickly rises between epochs 5 and 15, then stabilizes. This would indicate that the hyperparameter of 50 search epochs (also used by DARTS \cite{liu2019darts}) is way too large and could be reduced. 

This intuition is further confirmed by the fact that this model quickly converged to a very high top-1 accuracy (90 \%) around the same epoch (15) that the distance metric reached a plateau (see Fig. \ref{fig:val_acc_cifar10}). After epoch 20, the validation accuracy slowly converges towards 100 \%. It is most likely due to the optimization of network weights rather than architectural modifications, since $M$ remains stable.

Following the facts presented above, we decided to launch a new optimization of ResNet18 on CIFAR10 with the number of epochs reduced from 50 to 15. We trained the models obtained from the 50-epoch and 15-epoch optimizations for 600 epochs. As shown in Table \ref{tab:tab_metric}, the performance level is marginally degraded, while still being a considerable improvement over the original ResNet18 implementation (93.75\%, see Table \ref{tab:cifar10}). However, the search cost diminished significantly from 0.3 GPU-day to 0.1 GPU-day. This highlights the usefulness of our novel distance metric $M$.

\begin{table}[ht]
    \centering
    \caption{Results of training for 600 epochs two DARTOpti models optimized from ResNet18 \cite{he2016deep} on CIFAR-10 \cite{krizhevsky2009learning} with different numbers of search epochs.}
    \begin{tabular}{lllll}
        \toprule
        \thead{Starting\\ Architecture} & \thead{Search\\ Epochs} & \thead{Train\\ Epochs}  & \thead{ Validation \\ Top-1 (\%)} & \thead{Search Cost\\ (GPU-days)}\\
        \midrule
        ResNet18 & 50 & 600 & 97.39 & 0.3\\
        ResNet18 & 15 & 600 & 96.93 & 0.1 \\
        \bottomrule
    \end{tabular}
    \label{tab:tab_metric}
\end{table}

Furthermore, inspired by this analysis, we designed and implemented a mechanism that will automatically end the search process if the distance metric remains stable for 5 consecutive epochs starting from epoch 10 (by taking into account the 5 pretraining epochs). Empirically, we observed that this mechanism activated more often when searching on CIFAR-10 than CIFAR-100 and never on ImageNet (as the latter is more challenging, and the model makes better use of the 50 pre-allocated search epochs).

\subsection{Searching Architectures on CIFAR}
\label{subsec:cifar}

When searching and evaluating on CIFAR-10 and CIFAR-100~\cite{krizhevsky2009learning}, we mainly use 8 and 14 layer networks and select either 1 or 2 operations per edge to test how smaller architectures compete with larger ones. All results are presented in detail in Table \ref{tab:cifar10} and Table \ref{tab:cifar100}. 

Overall, D-DARTS models reach competitive results in both datasets. The smaller ones, such as DD-1 or DD-5, can match the performance of previous baselines despite possessing fewer parameters (e.g., 1.7M against 2.8M for the smallest model of \cite{chu2020fair}), although the largest achieve better results (e.g., $84.15\%$ top-1 accuracy for DD-4 on CIFAR-100). 

Moreover, Algorithm \ref{algo:arch_deriv} effectively provides a performance gain in both datasets (e.g., around $0.3\%$ for DD-4 when using 14 cells instead of 8), thus asserting its usefulness. We also compared the performance of models using FairDARTS~\cite{chu2020fair} loss function $L_F$ (see Eq. (\ref{eq:fair_darts_loss})) with ones using our new ablation-based total loss function $L_{T}$ (see Eq. (\ref{eq:total_loss})) to assert its effectiveness (see Section \ref{sec:ablation_loss_impact} for details).

Furthermore, there is a consequent performance variation when using the \textit{edge} parsing method rather than the \textit{sparse} one (e.g., around $0.5\%$ between DD-1 and DD-4 on CIFAR-10) while making the number of parameters double. Interestingly, this impact is more important on CIFAR-100 (around $2\%$). This may be related to the fact that using deeper architectures is less relevant with simpler datasets such as CIFAR-10, where \textit{sparse} models already achieve very high top-1 scores (greater or equal to $97\%$), than with more challenging datasets like CIFAR-100 or ImageNet~\cite{krizhevsky2012imagenet}. 

When considering the optimization of handcrafted architectures, Table~\ref{tab:cifar10} shows that DARTOpti (introduced in Section \ref{sec:encoding_darts}) managed to improve the performance of three architectures (ResNet18, ResNet50~\cite{he2016deep}, and Xception~\cite{chollet2017xception}). Top-1 accuracy was increased by an average of $2\%$ for CIFAR-10 and $3.1\%$ for CIFAR-100, making the optimized architectures comparable with other methods from the literature such as FairDARTS~\cite{chu2020fair} and the original DARTS~\cite{liu2019darts}. This performance rise is especially high for ResNet18 ($3.63\%$ for CIFAR-10, $3.3\%$ for CIFAR-100) and ResNet50 ($2.05\%$ for CIFAR-10, $5.5\%$ for CIFAR-100) but is less important for Xception ($0.3\%$ for CIFAR-10, $0.55\%$ for CIFAR-100). This discrepancy in performance gain could be explained by the fact that Xception is a much harder architecture to optimize than ResNets since the number of cells more than triples (i.e., going from 4 to 13).

To conclude, these results demonstrate the usefulness and relevance of our proposed approach.

\begin{table*}[ht]
    \begin{center}
    \caption{\label{tab:cifar10}Comparison of models on CIFAR-10 \cite{krizhevsky2009learning}. Each reported Top-1 accuracy is the best of 4 independent runs. For previous baselines, results are the official numbers from their respective articles. The search cost is expressed in GPU days. $\diamond$: Our implementation in DARTS-based search space $S_o$, results might vary from the official one.}
    \begin{tabular}{lllllllll}
        \toprule
        \thead{Models} & \thead{Params\\ (M)} & \thead{Parsing\\ Method} & \thead{Loss} & \thead{Top-1\\ (\%)} & \thead{Layers} & \thead{Cost} & \thead{Searched\\ On} & \thead{Type} \\
        \midrule
        NASNet-A\cite{zoph2017neural} & 3.3 & N.A. & N.A. & 97.35 & N.A. & 2000 & CIFAR-10 & RL\\
        DARTS\cite{liu2019darts} & 3.3 & \textit{darts} & $L_{CE}$ & 97.00 & 20 & 1.5 & CIFAR-10 & GD\\
        PC-DARTS\cite{xu2020pcdarts} & 3.6 & \textit{darts} & $L_{CE}$ & 97.43 & 20 & 3.8 & CIFAR-10 & GD\\
        P-DARTS\cite{chen2021progressive} & 3.4 & \textit{darts} & $L_{CE}$ & 97.50 & 20 & \textbf{0.3} & CIFAR-10 & GD\\
        FairDARTS-a\cite{chu2020fair} & 2.8 & \textit{sparse} & $L_F$ & 97.46 & 20 & 0.4 & CIFAR-10 & GD\\
        {DOTS\cite{gu2021dots}} & 3.5 & \textit{sparse} & $L_{CE}$ & 97.51 & 20 & \textbf{0.3} & CIFAR-10 & GD\\
        {DARTS-\cite{chu2021darts}} & 3.5 & \textit{sparse} & $L_{CE}$ & 97.41 & 20 & 0.4 & CIFAR-10 & GD\\
        {$\beta$-DARTS\cite{ye2022beta}} & 3.75 & \textit{sparse} & $L_{CE}$ & 97.47 & 20 & 0.4 & CIFAR-10 & GD\\
        ResNet18\cite{he2016deep}$^\diamond$ & 14.17 & N.A. & N.A. & 93.75 & 4 & N.A. & N.A. & manual\\
        ResNet50\cite{he2016deep}$^\diamond$ & 24.36 & N.A. & N.A. & 95.15 & 4 & N.A. & N.A. & manual\\
        Xception\cite{chollet2017xception}$^\diamond$ & 14.7 & N.A. & N.A. & 96.29 & 13 & N.A. & N.A. & manual\\
        \multirow{9}{*}{Ours} \hspace{0.5em} DD-1 & \textbf{1.7} & \textit{sparse} & $L_T$ & 97.02 & 8 & 0.5 & CIFAR-10 & GD \\
        \hspace{3em} DD-2 & 3.3 & \textit{edge} & $L_{F}$ & 97.10 & \textbf{8} & 0.5 & CIFAR-10 & GD\\
        \hspace{3em} DD-3 & 6.55 & \textit{edge} & $L_{T}$ & 97.58 & 14 & 0.5 & CIFAR-10 & GD \\
        \hspace{3em} DD-4 & 3.9 & \textit{edge} & $L_{T}$ & 97.48 & 8 & 0.5 & CIFAR-100 & GD \\
        \hspace{3em} DD-4 & 7.6 & \textit{edge} & $L_{T}$ & \textbf{97.75} & 14 & 0.5 & CIFAR-100 & GD \\
        \hspace{3em} DO-1-ResNet18 & 36.32 & \textit{sparse} & $L_{T}$ & 97.39 & 4 & \textbf{0.3} & CIFAR-10 & GD \\
        \hspace{3em} DO-1-ResNet50 & 71.23 & \textit{sparse} & $L_{T}$ & 97.20 & 4 & \textbf{0.3}
        & CIFAR-10 & GD \\
        \hspace{3em} DO-1-Xception & 61 & \textit{sparse} & $L_{T}$ & 97.00 & 13 & 1.5 & CIFAR-10 & GD \\
        \bottomrule
    \end{tabular}
    \end{center}
\end{table*}

\begin{table*}[ht]
    \begin{center}
    \caption{\label{tab:cifar100}Comparison of models on CIFAR-100~\cite{krizhevsky2009learning}. Each reported Top-1 accuracy is the best of 4 independent runs. For previous baselines, results are the official numbers from their respective articles. The search cost is expressed in GPU days. $\dagger$: Results obtained by running the code released by the authors. $\diamond$: Our implementation in DARTS-based search space $S_o$, results might vary from the official one.}
    \begin{tabular}{lllllllll}
        \toprule
        \thead{Models} & \thead{Params\\ (M)} & \thead{Parsing\\ Method} & \thead{Loss} & \thead{Top-1\\ (\%)} & \thead{Layers} & \thead{Cost} & \thead{Searched\\ On} & \thead{Type} \\
        \midrule
        DARTS\cite{liu2019darts} & 3.3 & \textit{darts} & $L_{CE}$ & 82.34 & 20 & 1.5 & CIFAR-100 & GD\\
        P-DARTS\cite{chen2021progressive} & 3.6 & \textit{darts} & $L_{CE}$ & 84.08 & 20 & \textbf{0.3} & CIFAR-100 & GD\\
        FairDARTS\cite{chu2020fair}$^\dagger$ & 3.5 & \textit{sparse} & $L_F$ & 83.80 & 20 & 0.4 & CIFAR-100 & GD\\
        {DOTS\cite{gu2021dots}} & 4.1 & \textit{sparse} & $L_{CE}$ & 83.52 & 20 & \textbf{0.3} & CIFAR-100 & GD\\
        {DARTS-\cite{chu2021darts}} & 3.4 & \textit{sparse} & $L_{CE}$ & 82.49 & 20 & 0.4 & CIFAR-10 & GD\\
        {$\beta$-DARTS\cite{ye2022beta}} & 3.83 & \textit{sparse} & $L_{CE}$ & 83.48 & 20 & 0.4 & CIFAR-10 & GD\\
        ResNet18\cite{he2016deep}$^\diamond$ & 14.17 & N.A. & N.A & 78.38 & \textbf{4} & N.A. & N.A. & manual\\
        ResNet50\cite{he2016deep}$^\diamond$ & 24.36 & N.A. & N.A & 78.10 & \textbf{4} & N.A. & N.A. & manual\\
        Xception\cite{chollet2017xception}$^\diamond$ & 14.7 & N.A. & N.A & 78.25 & 13 & N.A. & N.A. & manual\\
        \multirow{9}{*}{Ours} \hspace{0.5em} DD-1 & \textbf{1.7} & \textit{sparse} & $L_{T}$ & 81.10 & 8 & 0.5 & CIFAR-10 & GD\\
        \hspace{3em} DD-4 & 3.9 & \textit{edge} & $L_{T}$ & 83.86 & 8 & 0.5 & CIFAR-100 & GD\\
        \hspace{3em} DD-4 & 7.6 & \textit{edge} & $L_{T}$ & \textbf{84.15} & 14 & 0.5 & CIFAR-100 & GD\\
        \hspace{3em} DD-5 & \textbf{1.7} & \textit{sparse} & $L_{T}$ & 81.92 & 8 & 0.5 & CIFAR-100 & GD\\
        \hspace{3em} DD-6 & 3.3 & \textit{edge} & $L_F$ & 82.90 & \textbf{8}  & 0.5 & CIFAR-100 & GD\\
        \hspace{3em} DD-6 & 6.1 & \textit{edge} & $L_F$ & 84.06 & 14 & 0.5 & CIFAR-100 & GD\\
        \hspace{3em} DO-2-ResNet18 & 51.52 & \textit{sparse} & $L_{T}$ & 83.21 & \textbf{4} & \textbf{0.3} & CIFAR-100 & GD \\
        \hspace{3em} DO-2-ResNet50 & 70.44 & \textit{sparse} & $L_{T}$ & 83.60 & \textbf{4} & \textbf{0.3} & CIFAR-100 & GD \\
        \hspace{3em} DO-2-Xception & 61 & \textit{sparse} & $L_{T}$ & 82.30 & 13 & 2 & CIFAR-100 & GD \\
        \bottomrule
    \end{tabular}
    \end{center}
    
\end{table*}

\subsection{Searching and Transferring to ImageNet}
\label{subsec:imagenet}

In order to test our approaches on a more challenging dataset, we transferred our best models searched on CIFAR-100 \cite{krizhevsky2009learning} to ImageNet \cite{krizhevsky2012imagenet}. We also searched directly on ImageNet using the same training tricks and hyperparameters as the authors of DARTS \cite{liu2019darts}. We trained models using RTX 3090 GPUs, and we kept the same hyperparameters and tricks as in \cite{chu2020fair, liu2019darts}. Training a model for 250 epochs takes around 7 days on a single GPU. 

Table \ref{tab:imagenet} shows that model DD-7 (searched directly on ImageNet) reached a top-1 accuracy of 75.5 \%, outperforming PC-DARTS \cite{xu2020pcdarts} and P-DARTS \cite{chen2021progressive} by 0.6 \%. It is important to note that all of these approaches (and ours) use DARTS search space $S$ while FairDARTS \cite{chu2020fair} uses its own custom search space (mainly composed of inverted bottlenecks) as they argue that $S$ is too limited for ImageNet. Nevertheless, DD-7 still reached a near-identical score as FairDARTS-D despite using this simpler search space. 

In addition, the DARTOpti versions of ResNet18 and ResNet50 reached a top-1 accuracy of respectively 77.0 \% and 76.3 \%. They critically improve on the original architectures, increasing top-1 accuracy by an average of 5.1 \%. Notably, DO-2-ResNet50 achieves the same score as FBNetV2 \cite{wan2020fbnetv2} while requiring nearly a hundred times less search cost (i.e., 0.3 GPU days versus 25 GPU days) and not even having been searched directly on ImageNet but instead transferred from CIFAR-100. 

Finally, a notable gap (around 0.5 \%) between DD-4 (transferred from CIFAR-100) and DD-7 shows that searching directly on ImageNet significantly impacts performance.

\begin{table*}[ht!]
    \begin{center}
    \caption{\label{tab:imagenet}Comparison of models on ImageNet \cite{imagenet_cvpr09}. For previous baselines, results are the official numbers from their respective articles. The search cost is expressed in GPU days. $\diamond$: Our implementation in DARTS-based search space $S_o$, results might vary from the official one.}
    \begin{tabular}{lllllllllllr}
        \toprule
        \thead{Models} & \thead{Params\\ (M)} & \thead{$+\times$\\ (M)} & \thead{Parsing\\ Method} & \thead{Loss} & \thead{Top-1\\ (\%)} & \thead{Layers} & \thead{Cost} & \thead{Search\\ Space} & \thead{Searched\\ On} & \thead{Type} \\
        \midrule
        FBNetV2-L1\cite{wan2020fbnetv2} & 8.49 & 326 & N.A. & N.A. & \textbf{77.0} & N.A. & 25 & custom & ImageNet & GD\\
        NASNet-A\cite{zoph2017neural} & 5.3 & 564 & N.A. & N.A. & 74.0 & N.A. & 2000 & custom & ImageNet & RL\\
        DARTS\cite{liu2019darts} & 4.7 & 574 & \textit{darts} & $L_{CE}$ & 73.3 & 14 & 4 & $S$ & CIFAR-100 & GD\\
        PC-DARTS\cite{xu2020pcdarts} & 5.3 & 586 & \textit{darts} & $L_{CE}$ & 75.8 & 14 & 3.8 & $S$ & ImageNet & GD\\
        P-DARTS\cite{chen2021progressive} & 5.1 & 577 & \textit{darts} & $L_{CE}$ & 75.9 & 14 & \textbf{0.3} & $S$ & ImageNet & GD\\
        FairDARTS-D\cite{chu2020fair} & 4.3 & 440 & \textit{sparse} & $L_F$ & 75.6 & 20 & 3 & custom & ImageNet & GD\\
        {DOTS\cite{gu2021dots}} & 5.3 & 596 & \textit{sparse} & $L_{CE}$ & 76.0 & 20 & 1.3 & $S$ & ImageNet & GD\\
        {DARTS-\cite{chu2021darts}} & 4.9 & 467 & \textit{sparse} & $L_{CE}$ & 76.2 & 20 & 4.5 & $S$ & ImageNet & GD\\
        {$\beta$-DARTS\cite{ye2022beta}} & 5.4 & 597 & \textit{sparse} & $L_{CE}$ & 75.8 & 20 & 0.4 & $S$ & CIFAR-100 & GD\\
        ResNet18\cite{he2016deep}$^\diamond$ & 14.17 & 2720 & N.A. & N.A. & 69.2 & 4 & N.A. & N.A. & manual\\
        ResNet50\cite{he2016deep}$^\diamond$ & 24.36 & 4715 & N.A. & N.A. & 73.9 & 4 & N.A. & N.A. & manual\\
        Xception\cite{chollet2017xception}$^\diamond$ & 14.7 & 31865 & N.A. & N.A. & 74.1 & 13 & N.A. & N.A. & manual\\
        \multirow{4}{*}{Ours} \hspace{0.5em} DD-4 & 7.6 & 617 & \textit{sparse} & $L_T$ & 75.0 & 14 & 0.5 & $S$ & CIFAR-100 & GD\\
        \hspace{3em} DD-7 & 6.4 & 828 & \textit{edge} & $L_{T}$ & 75.5 & 8 & 3 & $S$ & ImageNet & GD\\
        \hspace{3em} DO-2-ResNet18 & 53.4 & 8619 & \textit{sparse} & $L_T$ & \textbf{77.0} & \textbf{4} &\textbf{0.3} & $S_o$ & CIFAR-100 & GD \\
        \hspace{3em} DO-2-ResNet50 & 73.23 & 10029 & \textit{sparse} & $L_T$ & 76.3 & \textbf{4} & \textbf{0.3} & $S_o$ & CIFAR-100 & GD \\
        \bottomrule
    \end{tabular}
    \end{center}
    
\end{table*}

\subsection{{Detecting objects on MS-COCO and Instance Segmentation on Cityscapes}}

{We transferred our best model trained on ImageNet (DO-2-ResNet18) to MS-COCO \cite{lin2014microsoft} in order to test our approach on other tasks than image classification. We used our model as the backbone of RetinaNet \cite{lin2017focal} and fine-tuned for 12 epochs with MMDetection \cite{mmdetection}, similarly as FairDARTS \cite{chu2020fair}. Table \ref{tab:coco} shows that our approach reached a box AP score of 34.2 \% hence outperforming DARTS- \cite{chu2021darts}, FairDARTS \cite{chu2020fair} and the original ResNet18 \cite{he2016deep}.} 

{We also performed instance segmentation on Cityscapes \cite{Cordts2016Cityscapes}, a dataset that focuses on semantic understanding of street scenes. We used DO-2-ResNet18 as the backbone of Mask R-CNN \cite{he2017mask} and compared it against other baselines (DARTS, FairDARTS, ResNet18). Table \ref{tab:cityscapes} features results similar to MS-COCO, with D-DARTS outperforming all the other approaches.}

{This way, the performance advantage of D-DARTS is confirmed when transferring to computer vision tasks other than image classification. }

\begin{table*}[ht!]
    \begin{center}
    \caption{\label{tab:coco}Comparison of backbone models for RetinaNet \cite{lin2017focal} on MS-COCO \cite{lin2014microsoft}.}
    \begin{tabular}{llllllll}
        \toprule
        \thead{Models} & \thead{$AP$} (\%) & \thead{$AP_{50}$} (\%) & \thead{$AP_{75}$} (\%) & \thead{$AP_s$} (\%) & \thead{$AP_m$} (\%) & \thead{$AP_l$} (\%) \\
        \midrule
        FairDARTS\cite{chu2020fair} & 31.9 & 51.9 & 33.0 & 17.4 & 35.3 & 43.0\\
        DARTS-\cite{chu2021darts} & 32.5 & \textbf{52.8} & 34.1 & 18.0 & 36.1 &  43.4\\
        ResNet18\cite{he2016deep} & 31.7 & 49.6 & 33.4 & 16.2 & 34.2 & 43.0\\
        \textbf{DO-2-ResNet18 (Ours)} & \textbf{34.2} & 52.1 & \textbf{36.6} & \textbf{19.1} &\textbf{38.3} & \textbf{45.3} \\
        \bottomrule
    \end{tabular}
    \end{center}
    
\end{table*}

\begin{table*}[ht!]
    \begin{center}
    \caption{\label{tab:cityscapes}Comparison of backbone models for Mask R-CNN \cite{he2017mask} on Cityscapes \cite{Cordts2016Cityscapes}.}
    \begin{tabular}{llllllll}
        \toprule
        \thead{Models} & \thead{$AP$} (\%) & \thead{$AP_{50}$} (\%) & \thead{$AP_{75}$} (\%) & \thead{$AP_s$} (\%) & \thead{$AP_m$} (\%) & \thead{$AP_l$} (\%) \\
        \midrule
        FairDARTS\cite{chu2020fair} & 41.1 & 69.3 & N.A. & 18.9 & 42.4 & 61.8\\
        DARTS-\cite{chu2021darts} & 41.7 & \textbf{70.4} & N.A. & 19.5 & 43.4 & 62.3\\\
        ResNet18\cite{he2016deep} & 40.9 & 66.2 & N.A. & 17.6 & 41.1 & 61.8\\
        \textbf{DO-2-ResNet18 (Ours)} & \textbf{44.0} & 69.6 & N.A. & \textbf{20.2} &\textbf{46.0} & \textbf{65.1} \\
        \bottomrule
    \end{tabular}
    \end{center}
    
\end{table*}

\subsection{Statistics on the Search Space}
\label{subsec:statistics}

The metric described in Subsection \ref{sec:distance_metric} allows us to quantify the distance between architectures and thus can also be employed to draw statistics on the metric space. This way, we can determine the average amount of changes made by DARTOpti to the starting architectures.

\begin{figure}[ht!]
    \centering
    \includegraphics[width=\linewidth]{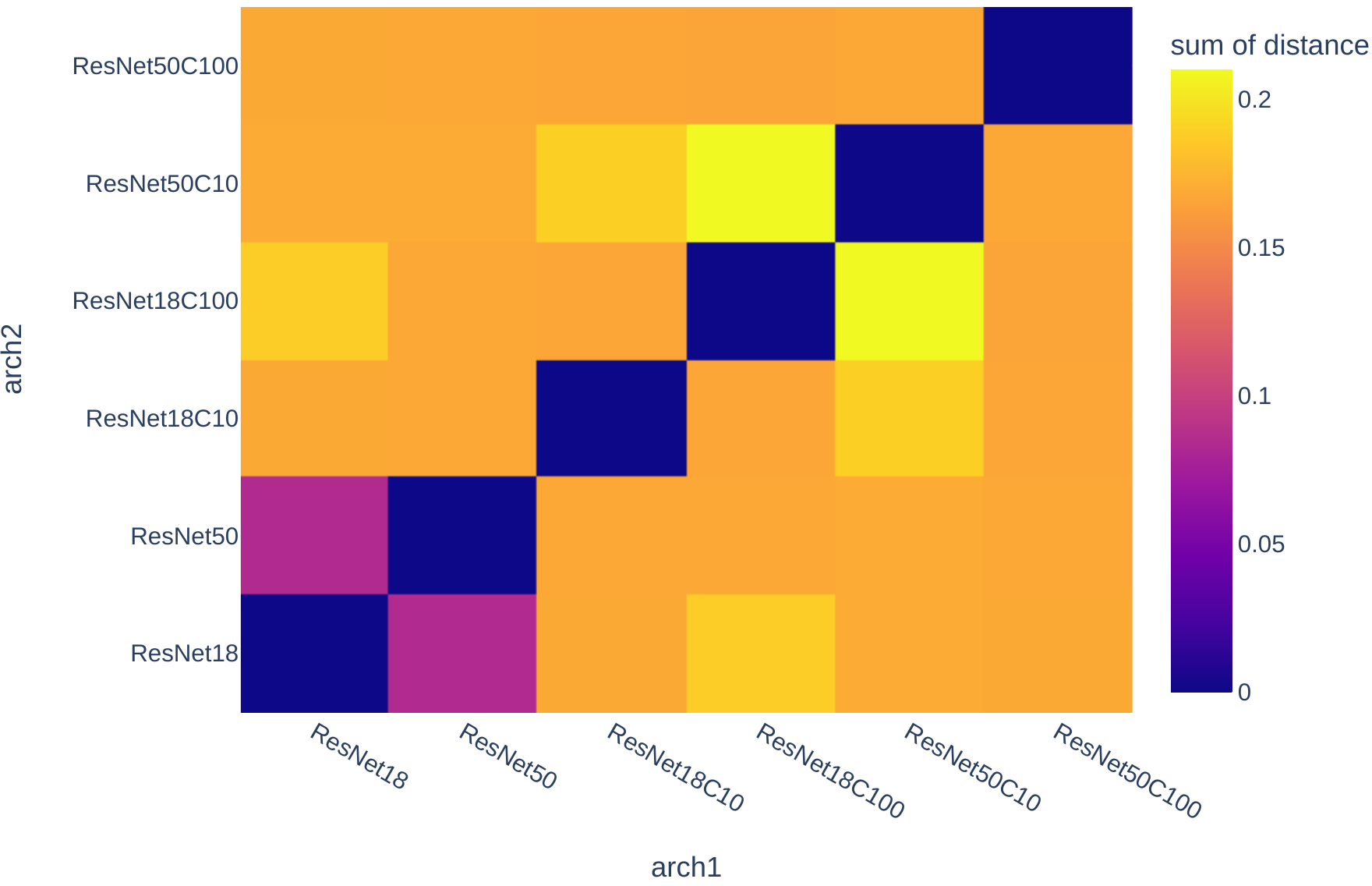}
    \caption{Heatmap representing the distance between the different architectures obtained from ResNet \cite{he2016deep} on CIFAR \cite{krizhevsky2009learning} datasets using the \textit{edge} parsing method.}
    \label{fig:distance_heatmap}
\end{figure}

Fig. \ref{fig:distance_heatmap} shows the distribution of changes between ResNet-based architectures. For instance, the distance between \textit{ResNet18C10} and \textit{ResNet50C10} is around 0.17 distance unit. The distance between architectures is relatively uniform, with the greatest (0.2 distance unit) being between \textit{ResNet18C100} and \textit{ResNet50C10}. This seems logical since they evolved from two different starting architectures (ResNet18 and ResNet50) and were also trained on two different datasets (CIFAR10 and CIFAR100). In addition, ResNet18 and ResNet50 are very close to each other as they both only consist of a few operations per cell (compared to DARTOpti architectures), and these operations are very similar (mostly skip connections and $3\times3$ convolutions). These observations indicate that both the starting architecture and the dataset used during the search process have an impact on the composition of the final architecture. 

\section{Discussion}
\label{sec:discussion}

In Section \ref{sec:approach}, we proposed D-DARTS, a new approach for differentiable architecture search based on a cell-level distributed search mechanism. Instead of searching for building blocks, we directly search for a complete super network composed of multiple subnets nested at the cell level. To bring an additional performance boost to this new mechanism, we also introduced an ablation-based loss function that leverages Game Theory in order to take into account the marginal contributions of each cell to the common goal (i.e., the classification task). Moreover, since it would be too costly to search for large networks directly, we presented Algorithm \ref{algo:arch_deriv}, a procedure to automatically derive larger architectures from smaller ones by stacking multiple times specific sequences of cells. In addition, we also described a way of leveraging our novel search process to directly optimize existing handcrafted architectures, denoted DARTOpti, along with a metric to quantify the distance between architectures.

In Section \ref{sec:experiments}, we showed that these proposed concepts perform well but are not exempt from limitations. {Increasing the search space size to such an extent (e.g., $10^{72}$ when considering 8 cells) makes the optimization process significantly more challenging. However, if we consider Fig. \ref{fig:convergence_speed}, we can observe that D-DARTS outperforms both DARTS and FairDARTS during the search phase. Hence, one interpretation of this fact is that expanding the search space provides benefits that far outweigh the increase in optimization difficulty. In addition, our novel ablation-based loss $L_{AB}$ (see Eq. \ref{eq:total_loss}) aims to enhance the optimization process by obtaining finer information (i.e., cell-specific) than the global Cross-Entropy loss $L_{CE}$. Algorithm \ref{algo:arch_deriv} also helps to reduce the optimization issue by allowing us to search on a small proxy network before expanding it to a larger one in the training phase.} In addition, D-DARTS is moderately less efficient w.r.t. memory (see Section \ref{sec:memory_efficiency}) than previous baselines \cite{liu2019darts, chu2020fair} due to the use of multiple optimizers. However, AMP substantially reduces memory usage, making D-DARTS more flexible. {Nevertheless, this optimization issue could potentially be further relieved by, for instance, enhancing cell optimizers. This could be the subject of future work.}

Furthermore, we proved that DARTOpti can successfully optimize top-performing handcrafted architectures such as ResNet50 \cite{he2016deep} with a significant gain in performance (i.e., a 3.9 \% average increase in top-1 accuracy across all datasets). However, this approach also has its limitations as the optimization process becomes more challenging when the architecture has many cells (e.g., Xception \cite{chollet2017xception} with 13 cells). This translates to an increased search cost and limited performance gains, as the standard 50 search epochs may not be enough for architectures of that size.

Nonetheless, combining the \textit{sparse} parsing method with our distributed design allows us to obtain unprecedentedly small architectures (around 1.7 M for the tiniest) that can still yield competitive results. In addition, while using the \textit{edge} parsing method, it is possible to search for larger size models that reach better results. This demonstrates the flexibility and usefulness of our novel approach.

\section{Conclusion and Future Work}
\label{sec:conclusion}

In this article, we proposed a novel paradigm for DARTS \cite{liu2019darts} with individualized cells. We showed that it effectively achieves competitive results on popular computer vision datasets for various tasks {(i.e., image classification, object detection, instance segmentation)}. We also demonstrated that this approach could successfully improve the design of existing top-performing handcrafted deep neural network architectures such as ResNet 50 \cite{he2016deep} or Xception \cite{chollet2017xception}. In addition, our novel architecture derivation algorithm allows us to build larger architectures from only a small number of cells without further training. 

However, this gain in performance does come at the cost of increased memory usage, which we alleviated by optimizing the search process and using AMP. Moreover, optimized architectures (DARTOpti) often see their number of parameters augment significantly, resulting in large architectures. Despite these downsides, we reached the same level of performance or better than previous baselines \cite{wan2020fbnetv2, chu2020fair} at a lower search cost. 

Finally, many ideas around this approach have not been explored yet, such as the automatic selection of search hyperparameters or enhancing cell optimizers. These could be the subject of future work.

\section*{Acknowledgments}

This work was performed using HPC resources from GENCI-IDRIS (Grant 20XX-AD011012644).

\printbibliography

\end{document}